\documentclass[pdflatex,sn-basic,Numbered]{sn-jnl}
\usepackage{graphicx}
\usepackage{amsmath,amssymb}
\usepackage{algorithm}
\usepackage{algpseudocode}
\usepackage{booktabs}
\usepackage{multirow}
\usepackage{xcolor}
\usepackage{pifont}
\usepackage{enumitem}
\usepackage{adjustbox}
\usepackage{geometry}
\usepackage{natbib}
\usepackage{capt-of} % allow captionof for side-by-side numbered figures
\usepackage{placeins} % for \FloatBarrier to control float placement
\usepackage{float} % for [H] exact figure placement (appendix layout control)
\newcommand{\safeinclude}[2][]{\IfFileExists{#2}{\includegraphics[#1]{#2}}{\fbox{\rule{0pt}{2.2in}\rule{0.95\linewidth}{0pt}}}}
\begin{document}

\title[XY-Cut++]{XY-Cut++: Advanced Layout Ordering via Hierarchical Mask Mechanism on a Novel Benchmark}

% Author and affiliation mapping
\author[1]{\fnm{Shuai} \sur{Liu}}\email{shuai\_liu@tju.edu.cn}
\author*[1]{\fnm{Youmeng} \sur{Li}}\email{liyoumeng@tju.edu.cn}
\author[1]{\fnm{Jizeng} \sur{Wei}}\email{weijizeng@tju.edu.cn}

\affil[1]{\orgname{College of Intelligence and Computing, Tianjin University}, \orgaddress{\city{Tianjin}, \postcode{300350}, \country{China}}}

\abstract{Document Reading Order Recovery is a fundamental task in document image understanding, playing a pivotal role in enhancing Retrieval-Augmented Generation (RAG) and serving as a critical preprocessing step for large language models (LLMs). Existing methods often struggle with complex layouts (e.g., multi-column newspapers), high-overhead interactions between cross-modal elements (visual regions and textual semantics), and a lack of robust block-level evaluation benchmarks. We introduce \textbf{XY-Cut++}, an advanced layout ordering method that integrates pre-mask processing, multi-granularity segmentation, and cross-modal matching to address these challenges. Our method significantly enhances layout ordering accuracy compared to traditional XY-Cut techniques. Specifically, XY-Cut++ achieves \textbf{state-of-the-art} performance (98.8 BLEU overall) while maintaining simplicity and efficiency. It outperforms existing baselines by up to 24\% and demonstrates consistent accuracy across simple and complex layouts on the newly introduced \textbf{DocBench-100} dataset. This advancement establishes a reliable foundation for document structure recovery, setting a new standard for layout ordering tasks and facilitating more effective RAG and LLM preprocessing. }

\keywords{Document reading order, Document layout analysis, XY-Cut++, DocBench-100}

\maketitle

% --- Teaser/overview figure ---
\begin{figure*}[t]
    \centering
    \safeinclude[width=\textwidth]{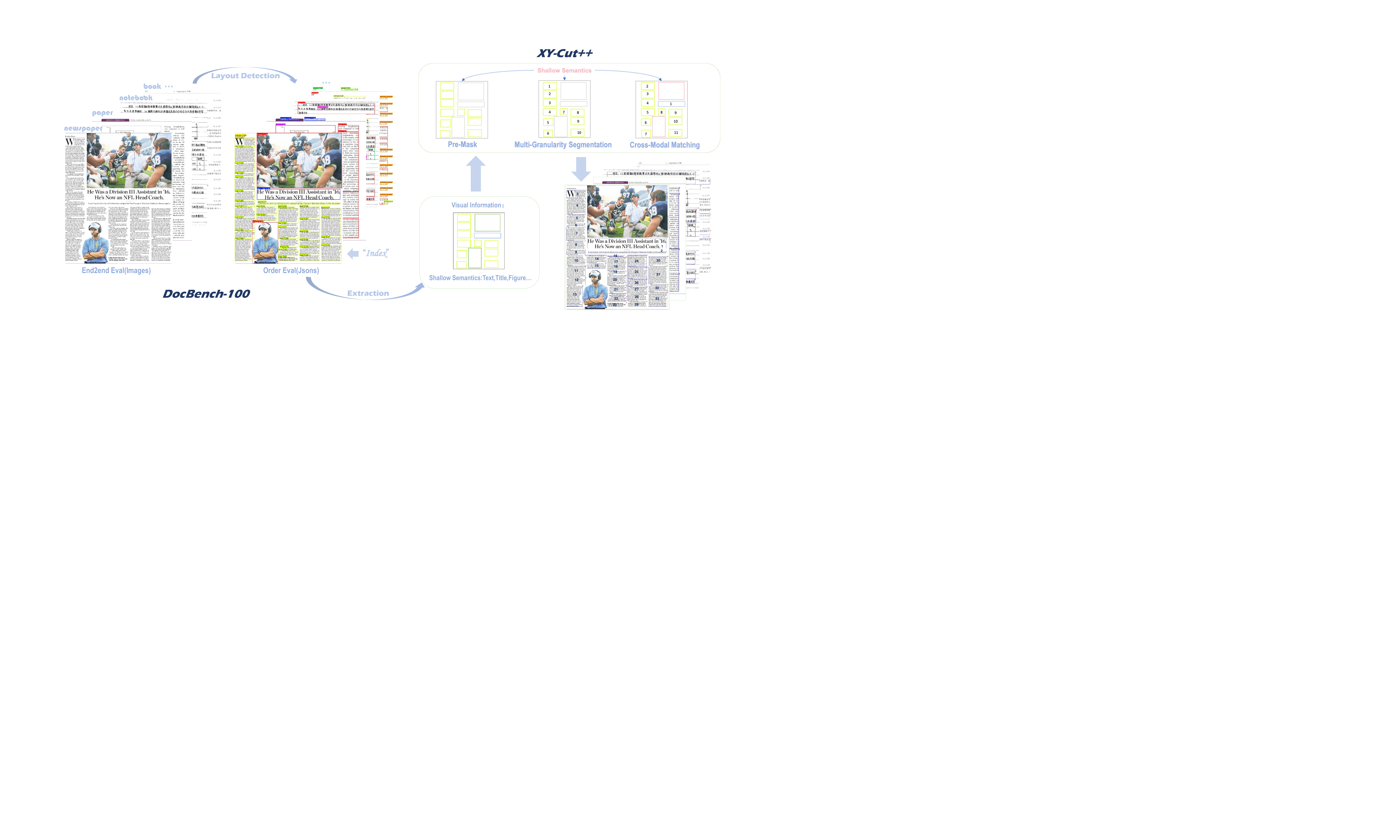}
    \caption{XY-Cut++ Architecture with Hierarchical Mask Mechanism and DocBench-100 Benchmark. (left) DocBench-100 dataset composition and dual evaluation protocols. (right) Algorithm workflow integrating adaptive pre-mask processing, multi-granularity segmentation, and cross-modal matching.}
    \label{fig:teaser}
\end{figure*}

\section{Introduction}\label{sec1}
\label{sec:intro}

Document Reading Order Recovery is a fundamental task in document image understanding, serving as a cornerstone for enhancing Retrieval-Augmented Generation (RAG) systems and enabling high-quality data preprocessing for large language models (LLMs). Accurate layout recovery is essential for applications such as digital publishing and knowledge base construction. However, this task faces several significant challenges: (1) complex layout structure (e.g., multi-column layout, nested text boxes, non-rectangular text regions, cross-page content), (2) inefficient cross-modal alignment due to the high computational costs of integrating visual and textual features, and (3) the lack of standardized evaluation protocols for block-level reading order. Traditional approaches like XY-Cut~\cite{xycut} fail to model semantic dependencies in complex designs, while deep learning methods such as LayoutReader~\cite{layoutreader} suffer from prohibitive latency, limiting real-world deployment. Compounding these issues, existing datasets like ReadingBank~\cite{layoutreader} focus on word-level sequence annotations, which inadequately evaluate block-level structural reasoning---a necessity for real-world layout recovery. Although OmniDocBench~\cite{omnidocbench} recently introduced block-level analysis support, its coverage of diverse complex layouts (e.g., newspapers, technical reports) remains sparse, further hindering systematic progress. Meanwhile, large-scale layout datasets such as DocBank~\cite{docbank} and PubLayNet~\cite{publaynet} have advanced document layout analysis but primarily target word/line-level or scientific article layouts and do not directly benchmark block-level reading order.

To address these challenges, we propose XY-Cut++, an advanced framework for layout ordering that incorporates three core innovations: (a) pre-mask processing to mitigate interference from high-dynamic-range elements (e.g., title, figure, table), (b) multi-granularity region splitting for the adaptive decomposition of diverse layouts, and (c) lightweight cross-modal matching leveraging minimal semantic cues. Our approach significantly outperforms traditional methods in layout ordering accuracy. Additionally, we introduce DocBench-100, a novel benchmark dataset designed for evaluating layout ordering techniques. It includes 100 pages (30 complex and 70 regular layouts) with block-level reading order annotations. Extensive experiments on DocBench-100 demonstrate that our method achieves state-of-the-art performance. Specifically, our method achieves BLEU-4 scores of 98.6 (complex) and 98.9 (regular), surpassing baselines by 24\% on average while maintaining 1.06$\times$ FPS of geometric-only approaches. 

The contributions of this paper are summarized as follows:
\begin{enumerate}
    \item We present a simple yet high-performance enhanced XY-Cut framework that fuses shallow semantics and geometry awareness to achieve accurate layout ordering. 
    \item We curate DocBench-100, a block-level benchmark with diverse complex layouts that complements existing datasets and directly targets reading-order evaluation. 
    \item We achieved state-of-the-art results on both existing and new block-level datasets, and provided extensive ablations to each component in our methods. 
\end{enumerate}

\begin{figure}[!b]
    \centering
    % Reduced to 0.8\linewidth to avoid occupying full width in single-column layout
    \safeinclude[width=0.6\linewidth]{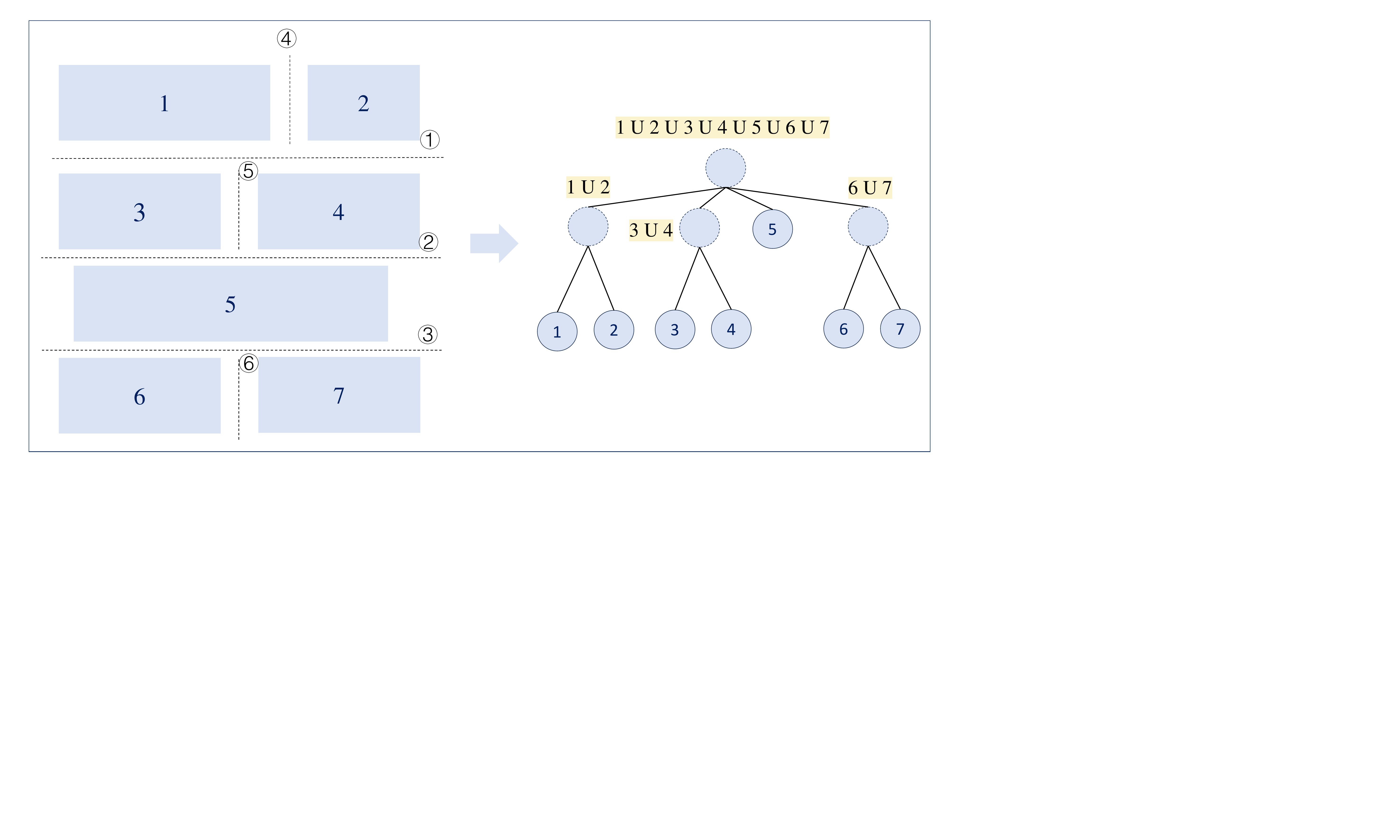}
    \caption{XY-Cut Recursive Partitioning Workflow and Failure Analysis in Complex Layouts: (1) Initial document segmentation steps, (2) Connectivity assumption violations caused by cross-layout cell structures (cell 5), and (3) Error propagation through subsequent layout ordering. The correct reading order is 1, 3, 2, 4, 5, 6, 7.}
    \label{fig:xycut}
\end{figure}

\section{Related Work}\label{sec2}
\label{sec:related}

Document Reading Order Recovery has advanced from early heuristic rule-based methods to modern deep learning models, yet reliably handling complex layouts remains challenging.

\subsection{Traditional Approaches}
\begin{figure}
    \centering
    \safeinclude[width=0.5\linewidth]{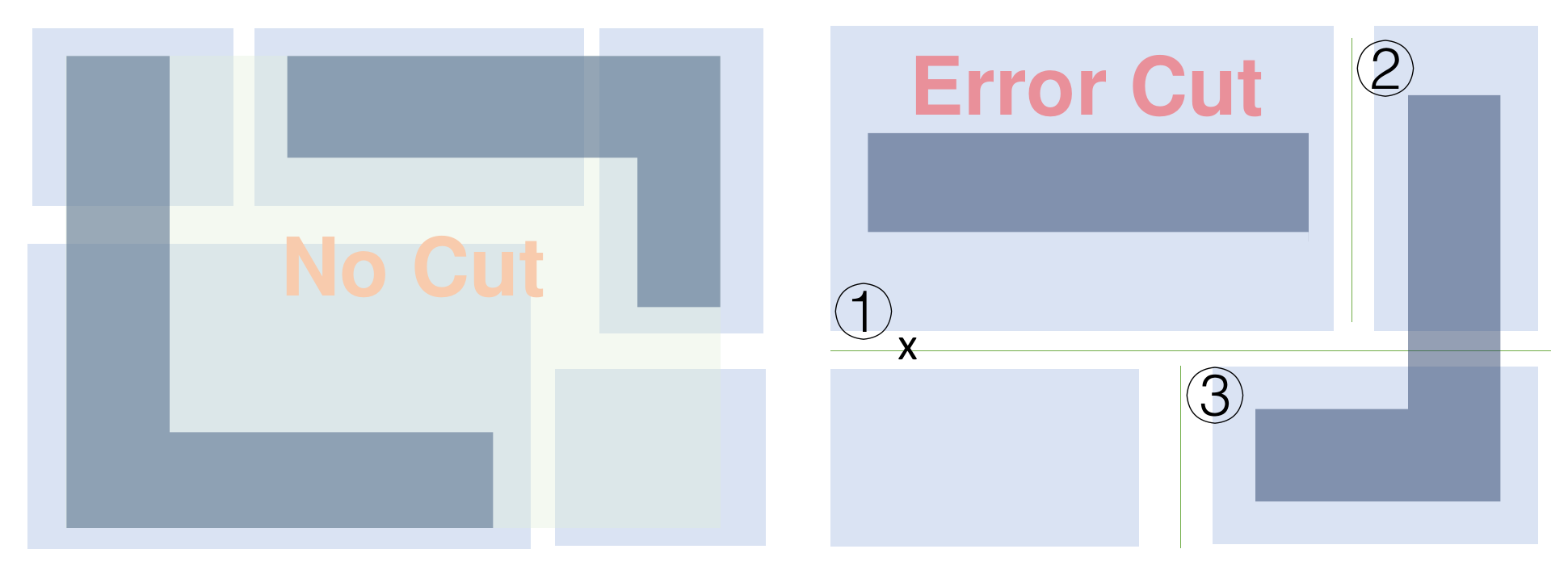}
    \caption{Challenges posed by L-shaped inputs: (1) segmentation failure due to the inability to process L-shaped structures, and (2) missegmentation caused by improper handling of L-shaped regions. The correct sequence of segmentation is \ding{193}+\ding{194} \ding{192}.}
    \label{fig:l-shape}
\end{figure}

The XY-Cut algorithm~\cite{xycut} is a foundational technique in Document Reading Order Recovery that recursively divides documents into smaller regions based on horizontal and vertical projections. As illustrated in Figure~\ref{fig:xycut}, while it is effective for simple layouts, it struggles with complex structures, leading to inaccuracies in reading order recovery. Specifically, the rigid threshold mechanisms of XY-Cut can introduce hierarchical errors when handling intricate layouts, resulting in suboptimal performance. 
To address these limitations, various enhancements have been proposed. For instance, dynamic programming optimizations~\cite{xycut_dyn} have been introduced to improve segmentation stability. Additionally, mask-based normalization techniques~\cite{xycut_mask} have been developed to mitigate some of the challenges associated with complex layouts. However, these improvements are still insufficient for handling intricate layouts and cross-page content. 
In our analysis of the XY-Cut algorithm, as discussed in~\cite{xycut_dyn,xycut_mask} and illustrated in Figure~\ref{fig:l-shape}, we identified that many challenging cases arise from L-shaped inputs. A straightforward solution involves initially masking these L-shaped regions and subsequently restoring them. Upon implementing this approach, we found it to be remarkably effective while maintaining both simplicity and efficiency. 

\subsection{Deep Learning Approaches}
Recent advances in Document Reading Order Recovery have been driven by deep learning models that effectively leverage multimodal cues. The LayoutLM series~\cite{layoutlm, layoutlmv2, layoutlmv3, layoutxlm, xylayoutlm} pioneered the integration of textual semantics with 2D positional encoding, where LayoutLMv2~\cite{layoutlmv2} introduced spatially aware pre-training objectives for cross-modal alignment and LayoutLMv3~\cite{layoutlmv3} further unified text/image embeddings through self-supervised masked language modeling and image-text matching. LayoutXLM~\cite{layoutxlm} extended this framework to multilingual document understanding. XYLayoutLM~\cite{xylayoutlm} advanced the field with its Augmented XY-Cut algorithm and Dilated Conditional Position Encoding, addressing variable-length layout modeling and generating reading order-aware representations for enhanced document understanding. Building upon these foundations, LayoutReader~\cite{layoutreader} demonstrated the effectiveness of deep learning in explicit reading order prediction through the sequential modeling of paragraph relationships. Critical to these advancements are large-scale datasets like DocBank~\cite{docbank}, which provides weakly supervised layout annotations for fine-grained spatial modeling, and PubLayNet~\cite{publaynet}, which contains over 360,000 scientific document pages with hierarchical labels that encode implicit reading order priors through structural patterns. Specialized benchmarks like TableBank~\cite{tablebank} further address domain-specific challenges by preserving the ordering of tabular data for table structure recognition. 

Architectural innovations have significantly enhanced spatial reasoning capabilities. DocRes~\cite{docres} introduces dynamic task-specific prompts (DTSPrompt), enabling the model to perform various document restoration tasks, such as dewarping, deshadowing, and appearance enhancement, meanwhile improving the overall readability and structure of documents. ~\cite{dla-backbone} iteratively aggregates features across different layers and resolutions, resulting in more refined feature representations that are beneficial for complex layout analysis tasks. By leveraging Hierarchical Document Analysis (HDA), models can more effectively capture intricate structural relationships within documents, facilitating more accurate predictions of reading order. Furthermore, advancements in unsupervised learning methods, such as those employed in DocUNet~\cite{docunet}, have enabled more effective handling of document distortions, thereby enhancing OCR performance and layout analysis accuracy. DocFormer~\cite{docformer} unified text, visual, and layout embeddings through transformer fusion, improving contextual understanding for logical flow prediction. Complementary to these approaches, EAST~\cite{east} established robust text detection through geometry-aware frameworks, serving as a critical preprocessing step for element-level sequence derivation. 

\subsection{Benchmarks for Document Reading Order Recovery}

Document Reading Order Recovery has seen significant advancements with deep learning models like LayoutLM~\cite{layoutlm} and LayoutReader~\cite{layoutreader}, which integrate visual and textual features for tasks such as reading order prediction. However, existing methods face critical limitations in handling complex layouts (e.g., multi-column structures, cross-page content). A key challenge is the lack of benchmarks that directly evaluate block-level reading order, which is essential for applications like document digitization. While datasets such as ReadingBank~\cite{layoutreader} provide word-level annotations for predicting reading sequences, their design complicates the evaluation of methods focused on block-level performance. Specifically, word-level sequence annotations cannot simplify the assessment of dependencies between text blocks (e.g., the order of paragraphs or figures spanning multiple columns), which are essential for modeling complex layouts. Recently, OmniDocBench~\cite{omnidocbench} has supported block-level analysis but suffers from limited coverage of complex layouts and sparse representation of domain-specific layouts (e.g., newspapers). To address these gaps, we introduce DocBench-100, which offers a broader range of layout structures and explicit metrics for assessing reading order accuracy, thereby enabling robust benchmarking of layout recovery systems. 

\section{DocBench-100 Dataset}\label{sec3}
\label{sec:docbench}
To enable rigorous evaluation of block-level reading order, we construct DocBench-100, a curated benchmark emphasizing diverse real-world layouts. It complements existing datasets (ReadingBank~\cite{layoutreader}, OmniDocBench~\cite{omnidocbench}, DocBank~\cite{docbank}, PubLayNet~\cite{publaynet}) by focusing on page-level block ordering across complex structures.

\subsection{Sources and Composition}
\begin{figure}
    \centering
    \includegraphics[width=1\linewidth]{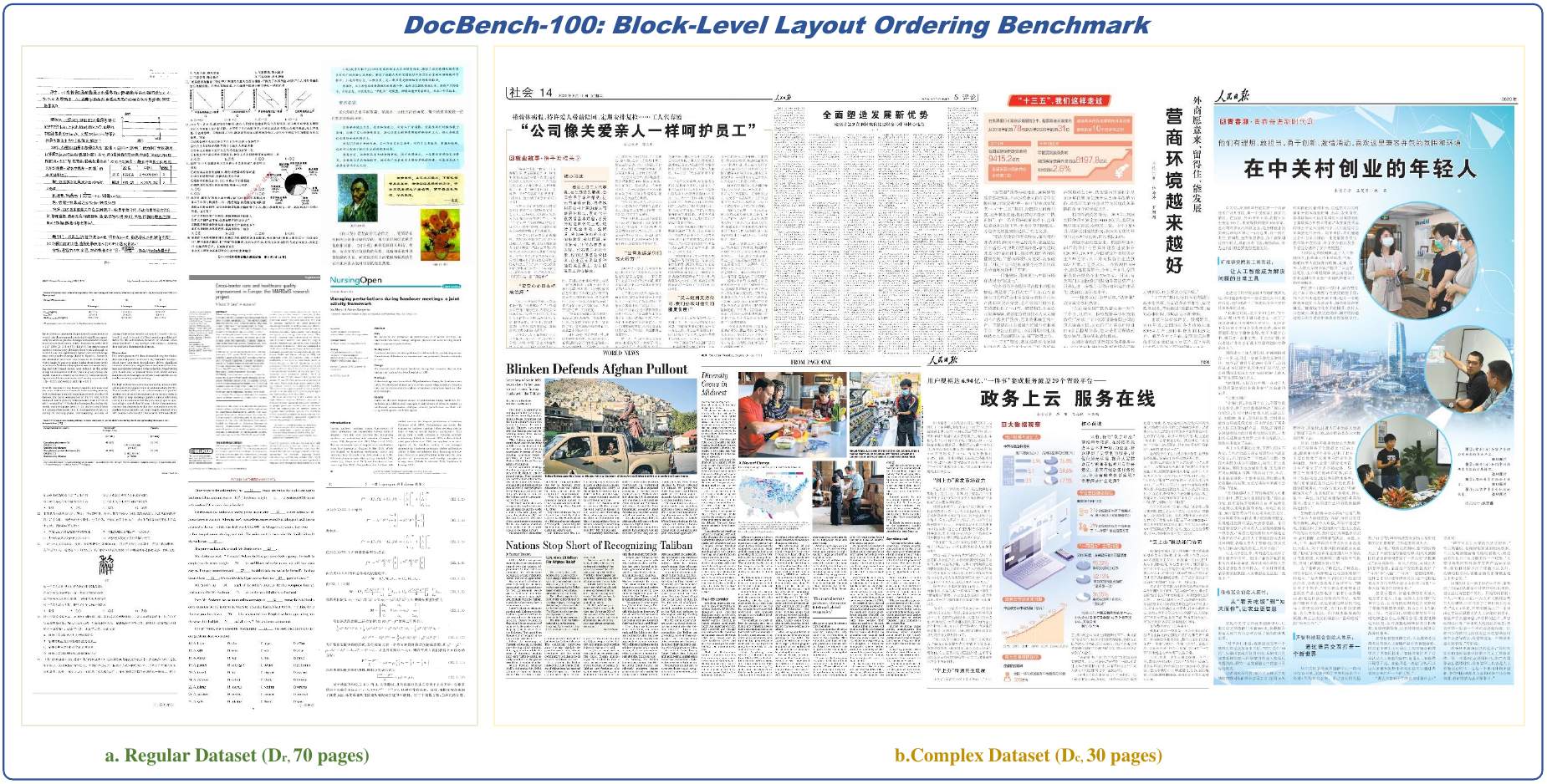}
    \caption{DocBench-100 image overview: (a) 70-page regular subset $D_r$, and (b) 30-page complex subset $D_c$. All pages provide block-level reading-order ground truth for benchmarking layout-ordering methods.}
    \label{fig:dataset}
\end{figure}

We source candidate pages from public document detection corpora (notably PP-DocLayout~\cite{pp-doclayout}) and MinerU~\cite{mineru} extraction outputs, then select pages exhibiting challenging phenomena: multi-column articles, irregular or spanning titles, interleaved figures/captions, and nested regions. The dataset comprises 100 pages split into: complex ($D_c$, 30 pages) and regular ($D_r$, 70 pages), following the prevalence in practical applications. An overview will be presented in Figure~\ref{fig:dataset}.

\subsection{Fields and File Structure}
Each page includes an image and two JSON files: (1) an input JSON (No Index) with fields \texttt{page\_id}, \texttt{page\_size}, \texttt{bbox}, \texttt{label}; (2) a GT JSON that additionally includes block-level reading order \texttt{index}. This design supports both end-to-end and oracle-detection evaluation protocols.

\subsection{Annotation Pipeline}
We adopt a two-stage pipeline:
\begin{itemize}[leftmargin=*,noitemsep]
    \item \textbf{Automatic pre-annotation}: MinerU~\cite{mineru} provides initial blocks, labels, and an order hypothesis.
    \item \textbf{Manual verification and screening}: Annotators correct segmentation and labels, and assign final \texttt{index}. Pages with inherently ambiguous global reading order (e.g., collage-like scans lacking semantic anchors) are excluded. All remaining reading orders are manually verified.
\end{itemize}

\subsection{Statistics and Usage Protocols}
$D_c$ contains predominantly multi-column layouts ($\geq 3$ columns) and irregular titles, while $D_r$ is dominated by single/double columns common in academic and business documents. Summary statistics by column structure are reported in Table~\ref{tab:docbench_stats}. We recommend reporting both: (a) end-to-end image-based evaluation (requires detection) and (b) JSON-based evaluation on block sequences. We will release the evaluation script for block-level BLEU and dataset documentation.

% --- Dataset statistics table ---
\begin{table}[t]
\caption{DocBench-100 subset statistics by column structure. Percentages are computed within each subset.}
\label{tab:docbench_stats}
\centering
\begin{tabular}{@{}lcccc@{}}
Subset & Pages & Single & Double & $\geq 3$ \\ 
\midrule
$D_c$ & 30 & 3.3\% & 6.7\% & 90.0\% \\
$D_r$ & 70 & 38.6\% & 54.4\% & 7.0\% \\
\bottomrule
\end{tabular}
\end{table}

% --- Placeholder for dataset overview figure ---
% \begin{figure*}[t]
%     \centering
%     \includegraphics[width=1\linewidth]{dataset.pdf}
%     \caption{DocBench-100 overview (placeholder). }
%     \label{fig:placeholder}
% \end{figure*}

\section{Methods}\label{sec4}
\label{sec:method}
\begin{figure*}[t]
    \centering
    \includegraphics[width=1\linewidth]{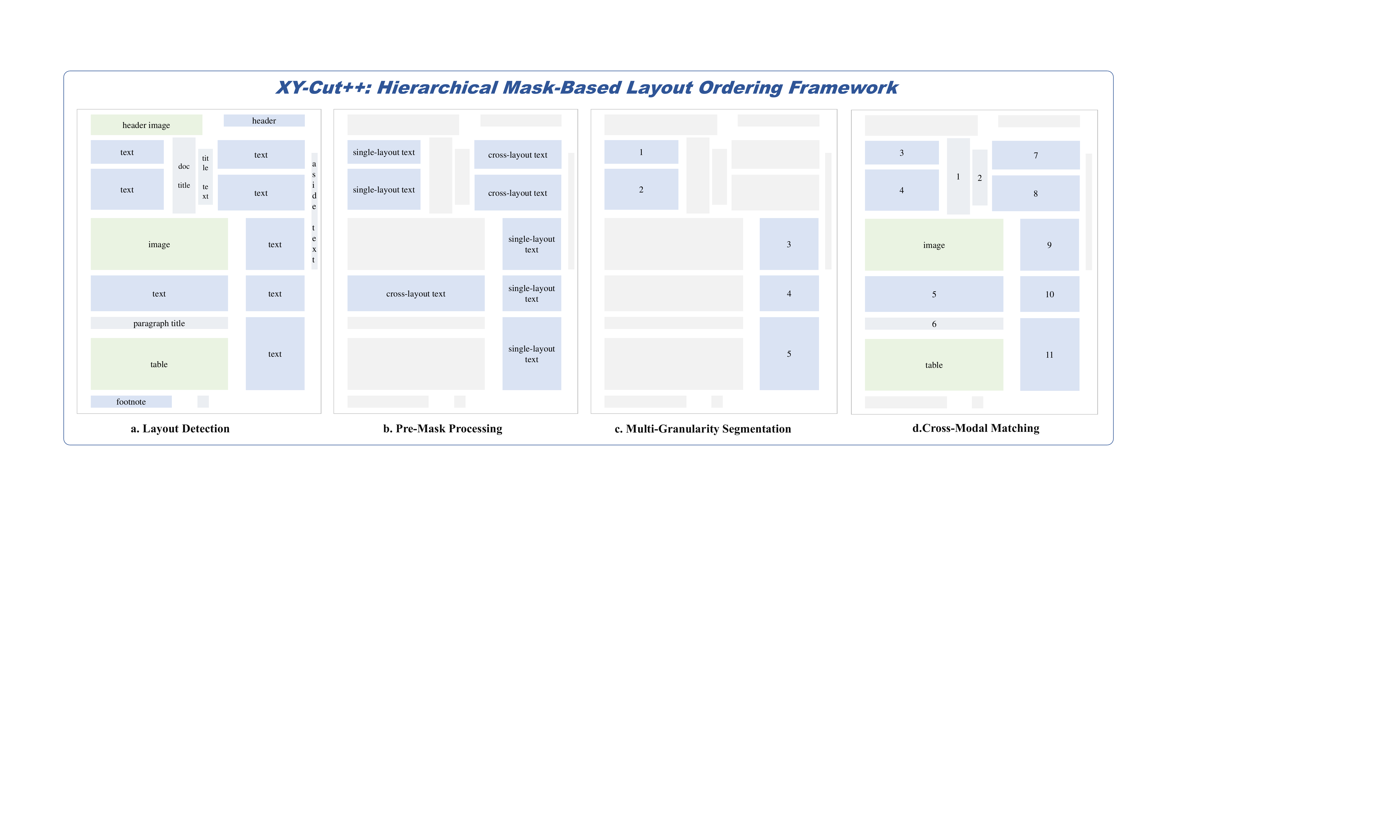}
    \caption{End-to-End Layout Ordering for Diverse Document Layouts Framework Overview: (a) Layout Detection (PP-DocLayout~\cite{pp-doclayout}), (b) Pre-Mask Processing, (c) Multi-Granularity Segmentation, and (d) Cross-Modal Matching. }
    \label{fig:overall}
\end{figure*}

Our geometry--semantic fusion pipeline (Figure~\ref{fig:overall}) has four stages: (1) PP-DocLayout~\cite{pp-doclayout} extracts visual features and shallow semantic labels; (2) highly dynamic elements (e.g., titles, tables, figures) are pre-masked to alleviate the L-shape problem; (3) cross-layout elements are identified via global analysis, then processed with masking, real-time density estimation, and heuristic sorting; and (4) masked elements are re-mapped using nearest IoU edge-weighted margins. This geometry-aware pipeline with shallow semantics attains state-of-the-art results on DocBench-100 and OmniDocBench~\cite{omnidocbench}.

\label{subsec:hierarchy}
\FloatBarrier % ensure preceding floats (e.g., Figure~\ref{fig:xycut}) are placed before the overview figure

% Place Figure~\ref{fig:mgs} and Figure~\ref{fig:cmm} on the same row while keeping separate numbering
\begin{figure}[t]
    \centering
    \begin{minipage}[t]{0.485\linewidth}
        \centering
        \includegraphics[width=\linewidth]{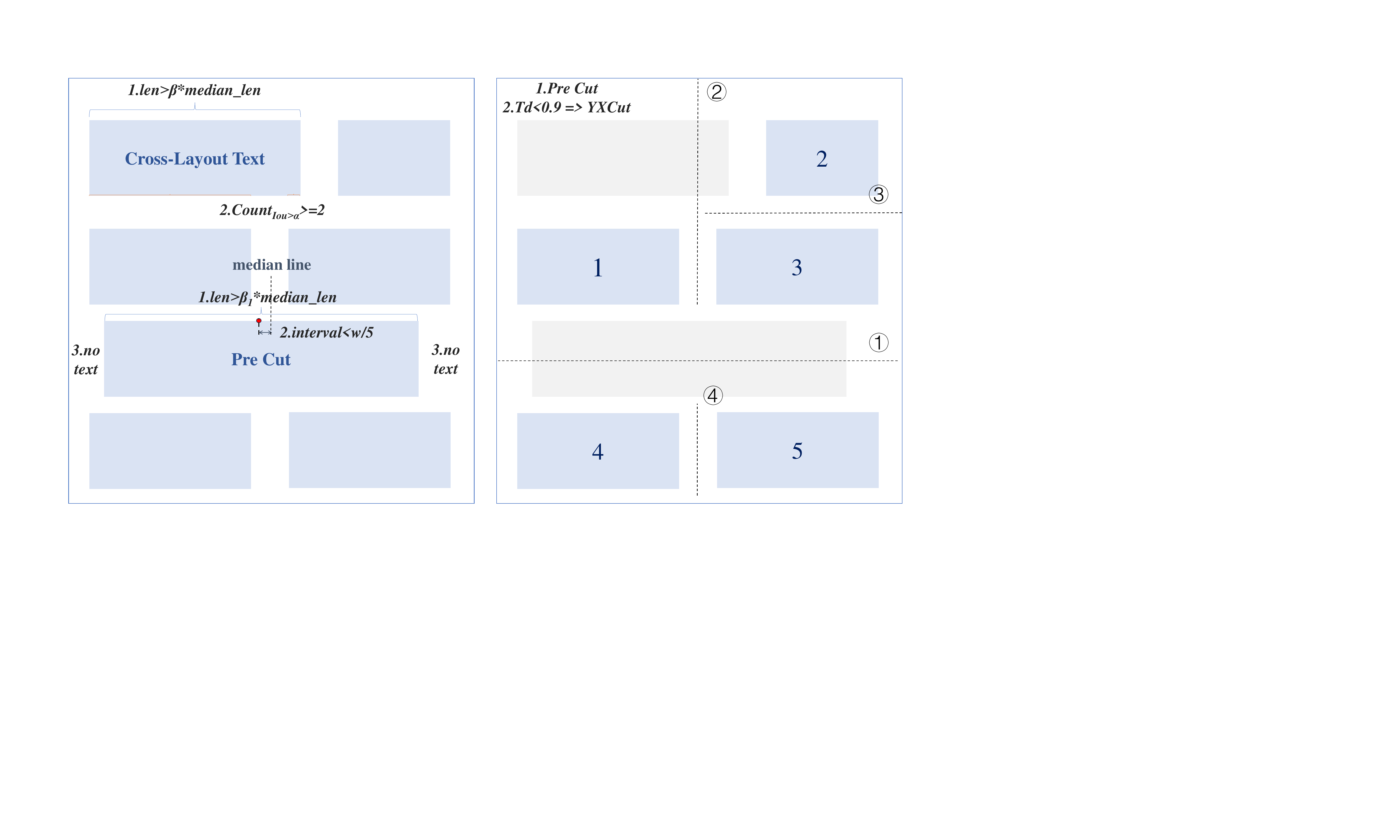}
    \captionof{figure}{Multi-Granularity Segmentation: (1) cross-layout element masking, (2) preliminary segmentation via pre-cut (\ding{172}), and (3) recursive density-driven partitioning. The enhanced XY-Cut algorithm adapts its splitting axis selection through real-time density evaluation ($\tau_d$), prioritizing horizontal splits for content-dense regions and vertical splits otherwise (\ding{173}\ding{174}). }
        \label{fig:mgs}
    \end{minipage}\hfill
    \begin{minipage}[t]{0.485\linewidth}
        \centering
        \includegraphics[width=\linewidth]{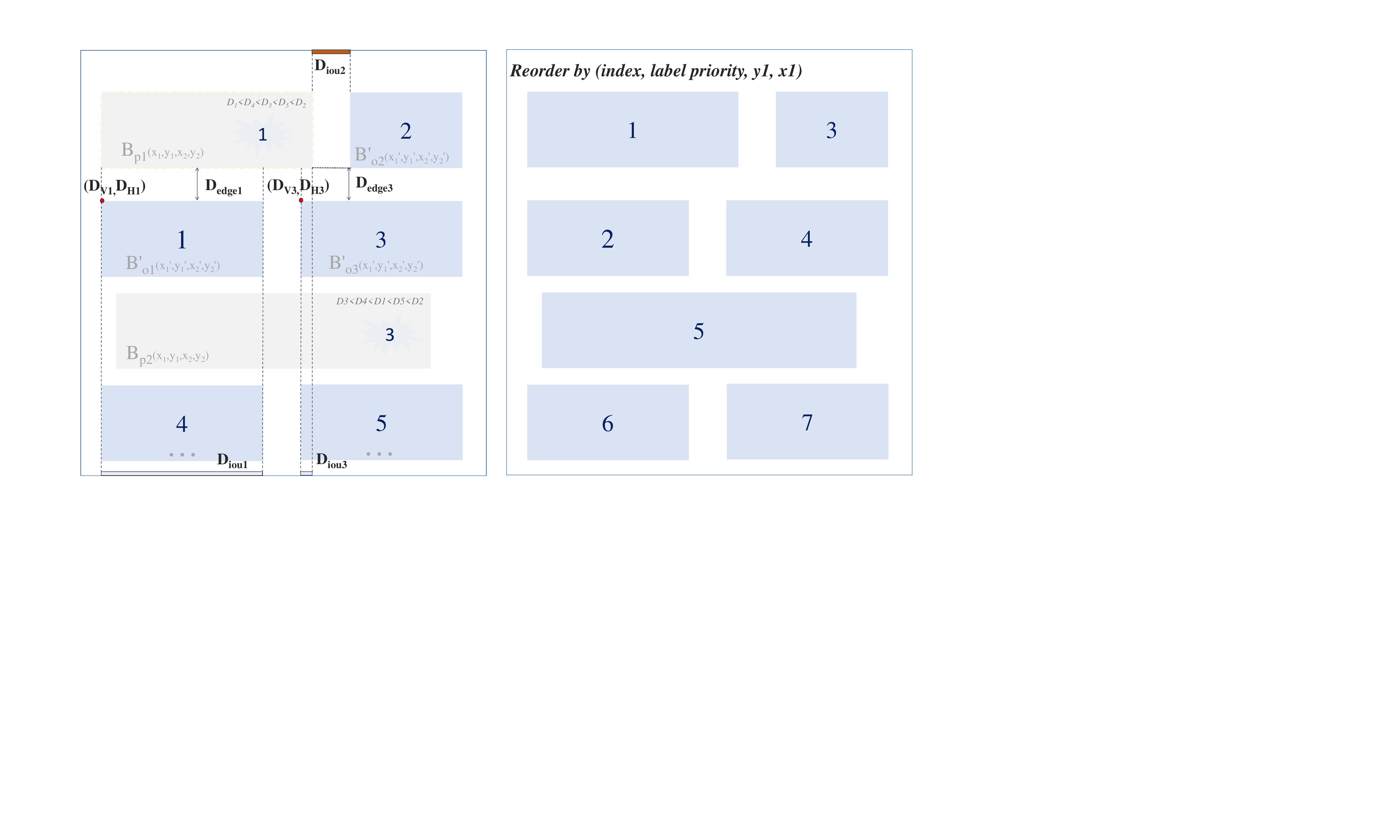}
        \captionof{figure}{Cross-Modal Matching: (1) semantic hierarchy-aware stage decomposition and (2) adaptive distance metric matching with dynamic policies and semantic-specific tuning. Subsequently, cells are reordered based on Index, Label Priority, Y1, and X1.}
        \label{fig:cmm}
    \end{minipage}
\end{figure}
\subsection{Pre-Mask Processing}
\label{subsec:premask}
Our Pre-Masking mechanism is designed to reduce the interference of highly dynamic elements (e.g., figures and tables) on core block sorting. These elements can appear anywhere in documents with high positional flexibility, which may disrupt sorting logic. Specifically, we first identify such elements using shallow semantic labels from the PP-DocLayout detection model. We then build a binary mask to temporarily exclude these elements from subsequent multi-granularity segmentation and core block sorting. After multi-granularity sorting, we remap the masked elements back to the ordered layout in the cross-modal matching phase, using an IoU-weighted distance nearest-neighbor strategy. This two-stage "mask-then-remap" pipeline effectively separates highly dynamic elements from the key sorting stage, ensuring accurate core block ordering while maintaining layout fidelity, and also reduces the previously mentioned "L"-shaped input issues.

\subsection{Multi-Granularity Segmentation}
\label{subsec:segmentation}
% As shown in Figure~\ref{fig:mgs}, our hybrid segmentation framework combines masking, preliminary splitting, and layout-aware refinement through three key phases:
As shown in Figure~\ref{fig:mgs}, our hybrid segmentation framework integrates cross-layout element masking, geometric pre-segmentation, and density-driven adaptive refinement in three key phases. Each phase targets specific layout challenges, enabling robust segmentation for both regular and complex documents.

\textbf{Phase 1: Cross-Layout Detection}
% Compute document-level median bounding box length and establish an adaptive threshold with scaling factor $\beta=1.3$: 
We first compute the document-level median bounding box length to establish an adaptive threshold for cross-layout elements (e.g., cross-column text spanning multiple grid units). Each content block \( B_i \) is represented by a 4-tuple \( (x_1, y_1, x_2, y_2) \), where \( (x_1, y_1) \) is the top-left corner and \( (x_2, y_2) \) is the bottom-right corner. Let \( \{l_i\}_{i=1}^N \) denote the length of each content bounding box \( B_i \): for horizontal-layout documents (commonly used in daily scenarios, e.g., reports, articles), \( l_i = x_{2,i} - x_{1,i} \) (horizontal width); for vertical-layout documents (e.g., classical Chinese texts), \( l_i = y_{2,i} - y_{1,i} \) (vertical height). \( \beta = 1.3 \) is an empirically tuned scaling factor. The adaptive threshold \( \mathcal{T}_l \) is:
\begin{equation}
    \mathcal{T}_l = \beta \cdot \mathrm{median}(\{l_i\}_{i=1}^N)
    \label{eq:threshold}
\end{equation}

% \noindent Detect cross-layout elements using dual criteria:
Cross-layout elements are detected via two criteria: (1) \( l_i > \mathcal{T}_l \), and (2) horizontal projection overlap with at least 2 other blocks. The judgment function \( \mathcal{C}_{\mathrm{cross}}(B_i) \) is defined as:
\begin{equation}
    \mathcal{C}_{\mathrm{cross}}(B_i) = 
    \begin{cases} 
        1 & l_i > \mathcal{T}_l \ \land\ \sum_{j\neq i}\mathbb{I}_{\mathrm{overlap}}(B_i,B_j) \geq 2 \\
        0 & \text{otherwise}
    \end{cases}
    \label{eq:crosslayout-criterion}
\end{equation}
Here, \( \mathbb{I}_{\mathrm{overlap}}(B_i,B_j) \) is an indicator function (1 for horizontal projection overlap, 0 otherwise). Detected cross-layout elements are masked for subsequent layout-aware splitting, while single-layout elements (within one grid unit) are processed immediately to avoid interference.

\textbf{Phase 2: Geometric Pre-Segmentation}
% Central elements and isolated graphical components are identified through:
This phase aims to identify central content elements, including body titles (e.g., chapter and section titles), and isolated graphical components (e.g., figures and tables). These elements are leveraged to perform coarse-grained partitioning of the original page into multiple non-overlapping sub-regions, denoted as \( R \). Each \( R \) is subsequently subjected to density-driven adaptive refinement in Phase 3 to achieve fine-grained segmentation. Such elements are classified based on geometric features, formalized by the judgment function \( \mathcal{P}(B_i) \):
\begin{equation}
\mathcal{P}(B_i) = \mathbb{I}\left(\frac{\|c_i - c_{\text{page}}\|_2}{d_{\text{page}}} \leq 0.2\right) \land (\phi_{\text{text}}(B_i) = \infty) ,
\label{eq:preseg}
\end{equation}
Where:

\( c_i = (cx_i, cy_i) \): Center of \( B_i \), computed as \( cx_i = \frac{x_{1,i}+x_{2,i}}{2} \), \( cy_i = \frac{y_{1,i}+y_{2,i}}{2} \);
              
\( c_{\text{page}} \): Page center coordinate;
              
\( \|c_i - c_{\text{page}}\|_2 \): Euclidean distance between \( B_i \) and page center;
              
\( d_{\text{page}} \): The normalization factor  $ d_{\text{page}} $  is determined by the block type:
$$
d_{\text{page}} =
\begin{cases}
W_{\text{page}}, & \text{if } r_i < 3, \\
H_{\text{page}}, & \text{if } r_i \geq 3,
\end{cases}
\quad \text{where} \quad
r_i = \frac{w_i}{h_i} = \frac{x_{2,i} - x_{1,i}}{y_{2,i} - y_{1,i}},
$$ and  $ W_{\text{page}} $ ,  $ H_{\text{page}} $  are the page width and height, respectively.
              
\( \mathbb{I}(\cdot) \): Indicator function (1 if condition holds, 0 otherwise);
              
\( \phi_{\text{text}}(B_i) \): Minimum Euclidean distance from \( B_i \) to the nearest text block; \( \phi_{\text{text}}(B_i) = \infty \) means \( B_i \) is isolated (no adjacent text);
              
Target elements \( B_i \): Body titles (chapter/section titles), visual components (e.g., Figure, Table), all with high layout flexibility.
          
\noindent Elements with \( \mathcal{P}(B_i) = 1 \) are marked as isolated and added to the mask set. Their coordinates are used to split the page into non-overlapping sub-regions \( R \), laying the groundwork for recursive refinement.

\textbf{Phase 3: Density-Driven Refinement} 
To achieve fine-grained segmentation, we apply the adaptive XY-Cut algorithm to each coarse-grained sub-region \( R\) from Phase 2. This algorithm dynamically selects the splitting axis (horizontal/vertical) based on regional content density of the current sub-region \( R \). The density metric \( \tau_d \) quantifies the ratio of cross-layout element area to single-layout element area within \( R \), calculated as:
\begin{equation}
    \tau_d = \frac{\sum_{k=1}^{N_c} w_k^{(C_c)}h_k^{(C_c)}}{\sum_{k=1}^{N_s}w_k^{(C_s)}h_k^{(C_s)}} ,
    \label{eq:tau-density}
\end{equation}
where \( C_c \) and \( C_s \) are cross-layout and single-layout element sets in \( R \); \( N_c \), \( N_s \) are their counts. \( w_k^{(C_c)}h_k^{(C_c)} \) and \( w_k^{(C_s)}h_k^{(C_s)} \) represent the area of the \( k \)-th element in each set, respectively. 
% \( C_c \) denotes the set of cross-layout elements in \( R \), \( C_s \) denotes the set of single-layout elements in \( R \); \( N_c \), \( N_s \) are the counts of cross-layout and single-layout elements in \( R \), respectively; \( w_k^{(C_c)} \), \( h_k^{(C_c)} \) are the width and height of the \( k \)-th cross-layout element, so \( w_k^{(C_c)}h_k^{(C_c)} \) is its area; \( w_k^{(C_s)} \), \( h_k^{(C_s)} \) are the width and height of the \( k \)-th single-layout element, and their product is its area.

A higher \( \tau_d \) indicates a denser distribution of cross-layout elements (typically horizontal content). We set a density threshold \( \theta_v = 0.9 \) to determine the splitting direction, as defined by the strategy function \( \mathcal{S}(R) \):
\begin{equation}
    \mathcal{S}(R) = 
    \begin{cases}
        \mathrm{XY\text{-}Cut} & \tau_d > \theta_v\  \\
        \mathrm{YX\text{-}Cut} & \text{otherwise}
    \end{cases}
    \label{eq:split}
\end{equation}
Here, XY-Cut refers to prioritizing horizontal splitting (along the Y-axis): we compute the Y-axis projection histogram of region \( R \), find the position \( y_{\text{cut}} \) with the minimum projection value (i.e., the "gap" between content blocks), and split \( R \) into upper and lower sub-regions. YX-Cut prioritizes vertical splitting (along the X-axis) using the same logic on the X-axis projection histogram.

The recursive splitting process terminates when each sub-region contains only one non-masked content block. Each final atomic region \( R_i \) is represented as a 7-tuple, encapsulating spatial coordinates, content type, semantic label, and positional index:
\begin{equation}
    R_i = \langle x_1^{(i)}, y_1^{(i)}, x_2^{(i)}, y_2^{(i)}, C_i, \text{Label}_i, \text{Index}_i \rangle
    \label{eq:region-repr}
\end{equation}
where \( (x_1^{(i)}, y_1^{(i)}) \) and \( (x_2^{(i)}, y_2^{(i)}) \) denote the top-left and bottom-right coordinates of \( R_i \); \( \mathcal{C}_{\mathrm{type}} = \{\text{cross-layout}, \text{single-layout}\} \), with \( C_i \in \mathcal{C}_{\mathrm{type}} \) indicating the content type of \( R_i \); \( \text{Label}_i \) represents the semantic label (e.g., title, figure, paragraph) of \( R_i \); and \( \text{Index}_i \) denotes the positional index of \( R_i \) in the document layout. This 7-tuple representation provides comprehensive information to support subsequent cross-modal matching.

\subsection{Cross-Modal Matching}
\label{subsec:matching}
To establish reading coherence consistent with human habits, we propose a geometry-aware cross-modal alignment framework, consisting of multi-stage semantic filtering and adaptive distance metric modules (Figure~\ref{fig:cmm}). This framework fuses semantic priority and geometric constraints to realize accurate ordering of atomic regions.

\textbf{Multi-Stage Semantic Filtering}: This module realizes the restoration of masked elements and optimizes the ordering sequence by leveraging label priority, ensuring high-semantic elements (e.g., cross-layout text, titles) are prioritized in the matching and restoration process. We first define core sets and a global label priority sequence, then formalize the matching logic to sequentially restore masked elements into the ordered sequence.

The multi-stage semantic filtering module restores masked elements by leveraging a global label priority sequence $\mathcal{L}_{\text{order}}$. The process operates on three core sets: the pre-ordered atomic region set $S$, the masked element set $\mathcal{M}$, and the dynamically updated target sequence $T$ (initialized as $T = S$). The mathematical formulation is:
\begin{equation}
\begin{gathered}
\mathcal{L}{_\text{order}}: \mathcal{L}{_\text{cross-layout}} \succ \mathcal{L}{_\text{title}} \succ \mathcal{L}{_\text{vision}} \succ \mathcal{L}{_\text{others}} \\
\mathcal{F}(B{_\text{p}}, T, l_{\text{current}}) = \begin{cases} 1, & \exists B_{\text{o}} \in T, \ \mathcal{L}{_\text{order}}(l(B{_\text{o}})) \succ \mathcal{L}{_\text{order}}(l_{\text{current}}) \\ 0, & \text{otherwise} \end{cases} \\
\mathcal{M}^{(l_{\text{current}})}{_\text{sorted}} = \{ (B{_\text{p}},B{_\text{best}}) \mid B{_\text{p}} \in \mathcal{M},\ l(B{_\text{p}})=l_{\text{current}},\ \mathcal{F}=1,\ B{_\text{best}} \in T \} \\
T = T \cup \mathcal{M}^{(l_{\text{current}})}{_\text{sorted}}, \quad \forall l_{\text{current}} \in \mathcal{L}{_\text{order}}
\end{gathered}
\label{eq:semantic-filtering}
\end{equation}
Here, the binary matching function $\mathcal{F}$ verifies for a pending element $B_{\text{p}}$ (with label $l_{\text{current}}$) whether a higher-priority candidate exists in $T$. For each $B_{\text{p}}$ that satisfies $\mathcal{F}=1$, the specific optimal anchor $B_{\text{best}} \in T$ is determined by minimizing a joint geometric distance (detailed in the following Adaptive Distance Metric). The set $\mathcal{M}^{(l_{\text{current}})}_{\text{sorted}}$ then contains all such matched pairs $(B_{\text{p}}, B_{\text{best}})$ for the current label. These pairs are subsequently integrated into $T$, and the process iterates through $\mathcal{L}_{\text{order}}$, ensuring high-priority elements are restored first.

This multi-stage semantic filtering strategy effectively eliminates irrational matching pairs (e.g., visual elements paired with footnotes) and reinforces the semantic coherence of the final ordered sequence, laying a solid foundation for downstream cross-modal alignment. 

\textbf{Adaptive Distance Metric}: We design a joint geometric distance metric with early termination to find the optimal matching position for each \( B_{\text{p}} \) in \( \mathcal{M}_{\text{sem}} \). Given a pending layout element \( B_{\text{p}} = (x_1,y_1,x_2,y_2) \) and an ordered candidate \( B_{\text{o}} = (x_1',y_1',x_2',y_2') \), the distance \( D(B_{\text{p}}, B_{\text{o}}, l) \) (parameterized by \( B_{\text{p}} \)'s label \( l \)) is a weighted sum of four geometric constraints \( \phi_k \):
\begin{equation}
D(B_{\text{p}}, B_{\text{o}}, l) = \sum_{k=1}^4 w_k \cdot \phi_k(B_{\text{p}}, B_{\text{o}}),
\label{eq:distance}
\end{equation}
where \( w_k \) are adaptive weights (detailed later), and \( \phi_1 \sim \phi_4 \) encode different geometric constraints to capture spatial relationships:

\begin{itemize}
    \item \textit{Intersection Constraints (\( \phi_1 \)): Filters invalid pairs by layout direction and overlap. The overlap threshold is set to \( \tau_{\text{overlap}} = 0.3 \). }
    \begin{equation}
    \phi_1 =
    \begin{cases}
    1, & \text{if } \mathrm{direction}(B_p) \neq \mathrm{direction}(B_o') \lor \mathrm{IoU}_{\text{projection}} < \tau_{\text{overlap}} \\
    0, & \text{otherwise}
    \end{cases}
    \label{eq:phi1}
    \end{equation}
    
    \item \textit{Boundary Proximity (\( \phi_2 \)): Measures spatial adjacency. Weighted by \( w_{\text{edge}} \), it uses center distances, prioritizing axis-aligned adjacency.}
    \begin{equation}
    \phi_2 = w_{\text{edge}} \times
    \begin{cases}
    d_x + d_y, & \text{(diagonal adjacency)} \\
    \min(d_x, d_y), & \text{(axis-aligned)}
    \end{cases}
    \label{eq:phi2}
    \end{equation}
    
    \item \textit{Vertical Continuity (\( \phi_3 \)): Optimizes vertical ordering. }
    \begin{equation}
    \phi_3 = \begin{cases} 
     -y_2', & l \in \mathcal{L}_{\text{cross-layout}} \land y_1 > y_2' \\
     y_1', & \text{(baseline alignment)}
    \end{cases}
    \label{eq:phi3}
    \end{equation}
    
    \item \textit{Horizontal Ordering (\( \phi_4 \)): Follows left-to-right reading logic via \( B_{\text{o}} \) left boundary.}
    \begin{equation}
    \phi_4 = x_1'
    \label{eq:phi4}
    \end{equation}
\end{itemize}

\noindent We introduce two parameterization strategies to optimize the metric's adaptability:

\textbf{Dynamic Weight Adaptation}. To ensure the four geometric constraint distances do not interfere with each other, we design scale-sensitive weights with staggered scaling levels. The weight vector \( \boldsymbol{w}_k = [w_1, w_2, w_3, w_4] \) is formulated as:
\begin{equation}
\boldsymbol{w}_k = [\max(h,w)^2, \max(h,w), 1, \max(h,w)^{-1}] 
\label{eq:dyn-weights}
\end{equation}
where h, w denote page dimensions, it establishes a distance metric with clear priorities. 

% Keep Figure~\ref{fig:complex} and Figure~\ref{fig:simple} on the same page (stacked) using a single float with two captionof blocks
\begin{figure}[t]
    \centering
    \begin{minipage}{1.0\linewidth}
        \centering
        \safeinclude[width=1\linewidth]{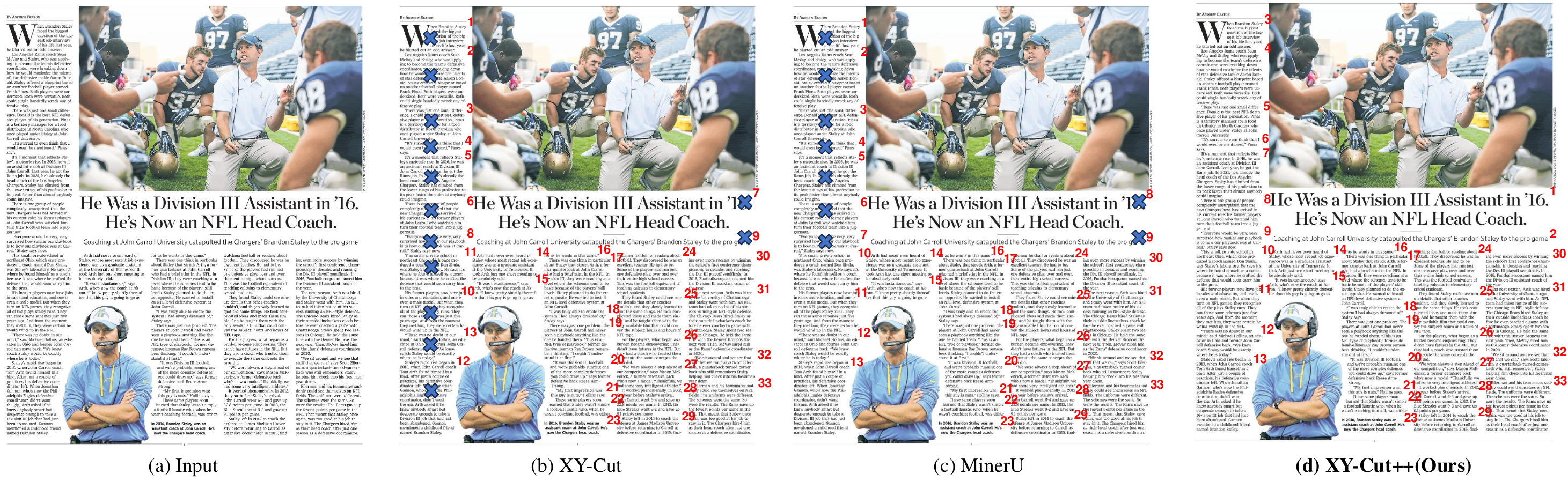}
        \captionof{figure}{Visualization of Complex Page \( D_c \) from DocBench-100 Dataset Using Different Layout Analysis Methods. \\
    (a) Input Image, (b) XY-Cut~\cite{xycut}, classic projection-based segmentation, (c) MinerU~\cite{mineru}, an end-to-end Document Content Extraction tool, (d) XY-Cut++, our proposed method. }
        \label{fig:complex}
    \end{minipage}
    \\\vspace{0.6em}
    \begin{minipage}{1.0\linewidth}
        \centering
        \safeinclude[width=1\linewidth]{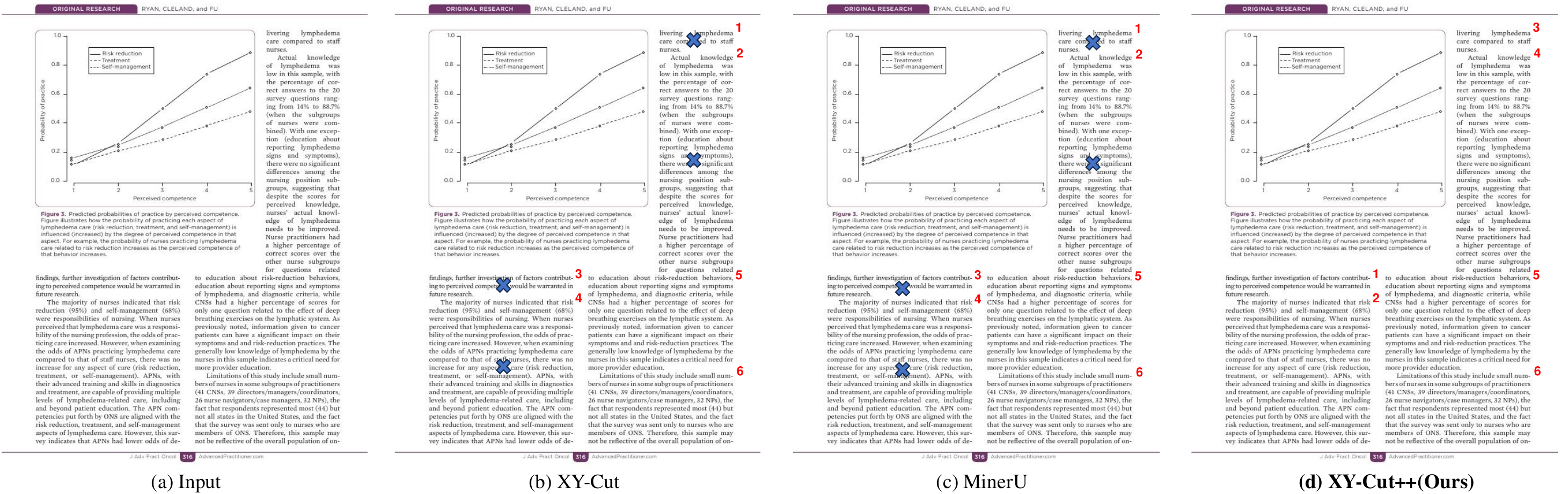}
        \captionof{figure}{Visualization of Regular Page \( D_r \) from DocBench-100 Dataset Using Different Layout Analysis Methods. \\
    (a) Input Image, (b) XY-Cut~\cite{xycut}, a classic projection-based segmentation, (c) MinerU~\cite{mineru}, an end-to-end Document Content Extraction tool, (d) XY-Cut++, our proposed method. }
        \label{fig:simple}
    \end{minipage}
\end{figure}

\textbf{Semantic-Specific Tuning}. Optimal weights from grid search on 2.8k documents:
\begin{equation}
\boldsymbol{w}_{\text{edge}} = \begin{cases}\label{eq:edge-weights}
[1,0.1,0.1,1] & l \in \mathcal{L}_{\text{title}} \cap \mathcal{O}_{\text{horizontal}} \\
[0.2,0.1,1,1] & l \in \mathcal{L}_{\text{title}} \cap \mathcal{O}_{\text{vertical}} \\
[1,1,0.1,1] & l \in \mathcal{L}_{\text{cross-layout}} \\
[1,1,1,0.1] & l \in \mathcal{L}_{\text{otherwise}}
\end{cases}
\end{equation}
where \( \mathcal{O}_{\text{horizontal}} \) and \( \mathcal{O}_{\text{vertical}} \) denote the horizontal layout set and vertical layout set, respectively; \( l \) is the semantic label of the element. This weight design assigns differentiated weights based on label characteristics and layout orientations, further weakening mutual interference between constraints and improving matching stability across various scenarios. 

Statistical validation demonstrates significant improvements: +2.3 BLEU-4 over uniform baselines. The overall process can be simply expressed in Algorithm~\ref{alg:cmm}. 

\begin{algorithm}[t]
\caption{Geometry-Aware Cross-Modal Matching with Semantic Filtering}
\begin{algorithmic}[1]
\Require Sorted anchor set $\mathcal{S}$, masked element set $\mathcal{M}$, distance weights $w_1, w_2, w_3, w_4$
\Ensure Matching result set $\mathcal{T}$
\State Initialize global label priority: $\mathcal{L}_{\text{order}}: \mathcal{L}_{\text{cross-layout}} \succ \mathcal{L}_{\text{title}} \succ \mathcal{L}_{\text{vision}} \succ \mathcal{L}_{\text{other}}$
\State $\mathcal{T} \gets \{ (B_o, B_o) \mid B_o \in \mathcal{S} \}$ \Comment{Initialize with self-pairs of anchors}

\For{each semantic label $l$ in $\mathcal{L}_{\text{order}}$ (descending priority)}
    \For{each pending box $B_p \in \mathcal{M}$ with label $l$} \Comment{Check $\mathcal{F}(B_p, \mathcal{T}, l)$}
        \State $D_{\min} \gets \infty$, $B_{\text{best}} \gets \text{null}$
        \For{each ordered box $B_o$ in $\mathcal{T}$} \Comment{Search candidates in $\mathcal{T}$ for $\mathcal{F}$}
            \State $D_{\text{curr}} \gets 0$
            \For{$k \gets 1$ to $4$}
                \State $D_{\text{curr}} \gets D_{\text{curr}} + w_k \cdot \phi_k(B_p, B_o)$
                \If{$D_{\text{curr}} > D_{\min}$}
                    \State \textbf{break} \Comment{Early termination}
                \EndIf
            \EndFor
            \If{$D_{\text{curr}} < D_{\min}$}
                \State $D_{\min} \gets D_{\text{curr}}$, $B_{\text{best}} \gets B_o$
            \EndIf
        \EndFor
        \If{$B_{\text{best}} \neq \text{null}$} \Comment{Valid match found for $B_p$}
            \State $\mathcal{T} \gets \mathcal{T} \cup \{(B_p, B_{\text{best}})\}$ \Comment{$\mathcal{M}_{\text{sorted}}^{(l)} \gets \mathcal{M}_{\text{sorted}}^{(l)} \cup \{(B_p, B_{\text{best}})\}$}
        \EndIf
    \EndFor
    \State Sort $\mathcal{T}$ by: Label priority of $B_p$ (desc) $\succ$ Index of $B_{\text{best}}$ (asc) $\succ$ $y_1$ of $B_p$ (asc) $\succ$ $x_1$ of $B_p$ (asc)
\EndFor
\State \Return $\mathcal{T}$ 
\end{algorithmic}
\label{alg:cmm}
\end{algorithm}

\section{Experiments}\label{sec5}
% (Removed standalone figure* for simple to keep both on one page in the combined float above)

\begin{table*}[t]
\caption{Progressive Component Analysis on DocBench-100.\textit{Metric Key:} BLEU-4 $\uparrow$ / ARD $\downarrow$ / Tau $\uparrow$. }
\label{tab:main}
\centering
\begin{tabular}{@{}lccccccccc@{}}
\toprule
Method & \multicolumn{3}{c}{$D_c$} & \multicolumn{3}{c}{$D_r$} & \multicolumn{3}{c}{$\mu$} \\
\cmidrule(lr){2-4} \cmidrule(lr){5-7} \cmidrule(lr){8-10}
 & BLEU-4& ARD & Tau & BLEU-4& ARD & Tau & BLEU-4& ARD  & Tau  \\
\midrule
XY-Cut & 0.749 & 0.233 & 0.878 & 0.819 & 0.098 & 0.912 & 0.797 & 0.139 & 0.902 \\
+Pre-Mask & 0.818 & 0.196 & 0.887 & 0.823 & 0.087 & 0.920 & 0.822 & 0.120 & 0.910 \\
+MGS & 0.946 & 0.164 & 0.969 & 0.969 & 0.036 & 0.985 & 0.962 & 0.074 & 0.980 \\
+CMM & \textbf{0.986} & \textbf{0.023} & \textbf{0.995} & \textbf{0.989} & \textbf{0.003} & \textbf{0.997} & \textbf{0.988} & \textbf{0.009} & \textbf{0.996} \\
\bottomrule
\end{tabular}
\end{table*}

\begin{table*}[h]
\caption{Ablation Study of Pre-Mask, Multi-Granularity Segmentation (MGS), and Cross-Modal Matching (CMM) on DocBench-100.\textit{Metric Key:} BLEU-4 $\uparrow$. }
\label{tab:mgs_cmm}
\centering
\begin{adjustbox}{max width=\textwidth}
\begin{tabular}{@{}lccccccccccc@{}}
\toprule
Method & Mask & Mask Cross-Layout & Pre-Cut & Adaptive Scheme & $\phi_1$ & $\phi_2$ & $\phi_3$ & $\phi_4$ & Dynamic Weights & Multi-Stage & BLEU-4 \\
\midrule
Baseline & &  &  &  &  &  &  &  &  &  & 0.797 \\
\hline
+Pre-Mask & \ding{51} &  &  &  &  &  &  &  &  &  & \textbf{0.822} \\
\hline
\multirow{4}{*}{+MGS} 
    & \ding{51} & \ding{51} &  &  &  &  &  &  &  &  & 0.905 \\
    & \ding{51} &  & \ding{51} &  &  &  &  &  &  &  & 0.914 \\
    & \ding{51} &  &  & \ding{51} &  &  &  &  &  &  & 0.923 \\
    & \ding{51} & \ding{51} & \ding{51} & \ding{51} &  &  &  &  &  &  & \textbf{0.962} \\
\hline
\multirow{7}{*}{+CMM} 
    & \ding{51} & \ding{51} & \ding{51} & \ding{51} & & & & & & \ding{51} & 0.963 \\
    & \ding{51} & \ding{51} & \ding{51} & \ding{51} & \ding{51} & & & & & \ding{51} & 0.765 \\
    & \ding{51} & \ding{51} & \ding{51} & \ding{51} & & \ding{51} & & &  & \ding{51} & 0.858 \\
    & \ding{51} & \ding{51} & \ding{51} & \ding{51} & & \ding{51} & & & \ding{51} & \ding{51} & 0.985 \\
    & \ding{51} & \ding{51} & \ding{51} & \ding{51} & & & \ding{51} & &  & \ding{51} & 0.881 \\
    & \ding{51} & \ding{51} & \ding{51} & \ding{51} & & & & \ding{51} &  & \ding{51} & 0.694 \\
    & \ding{51} & \ding{51} & \ding{51} & \ding{51} & \ding{51} & \ding{51} & \ding{51} & \ding{51} & \ding{51} & \ding{51} & \textbf{0.988} \\
\bottomrule
\end{tabular}
\end{adjustbox}
\end{table*}

\begin{table}[t]
\caption{Reading Order Recovery Performance: BLEU-4 $\uparrow$ Results on DocBench-100. The best results are in bold. }
\label{tab:all}
\centering
\begin{tabular}{@{}lccc@{}}
\toprule
Method & $D_c$ & $D_r$ & $\mu$ \\
\midrule
XY-Cut~\cite{xycut} & 0.749 & 0.818 & 0.797 \\
LayoutReader~\cite{layoutlmv3,layoutreader} & 0.656 & 0.844 & 0.788 \\
MinerU~\cite{mineru} & 0.701 & 0.946 & 0.873 \\
\midrule
\textbf{XY-Cut++(ours)} & \textbf{0.986} & \textbf{0.989} & \textbf{0.988} \\
\bottomrule
\end{tabular}
\end{table}

\begin{table*}[t]
\caption{Reading Order Recovery Performance on Textual Content of OmniDocBench (Excluding Figures/Tables with Insignificant Layout Impact). \textit{Metric Key:} BLEU-4 $\uparrow$ / ARD $\downarrow$ / Tau $\uparrow$. Best results are in \textbf{bold}; second-best are \underline{underlined}. }
\label{tab:omnidocbench}
\centering
\begin{adjustbox}{max width=\textwidth}
\begin{tabular}{@{}lccccccccccccccc@{}}
\toprule
Method & 
\multicolumn{3}{c}{Single} & 
\multicolumn{3}{c}{Double} & 
\multicolumn{3}{c}{Three} & 
\multicolumn{3}{c}{Complex} & 
\multicolumn{3}{c}{Mean} \\ 
\cmidrule(lr){2-4} \cmidrule(lr){5-7} \cmidrule(lr){8-10} \cmidrule(lr){11-13} \cmidrule(lr){14-16}
& {BLEU-4} & ARD & Tau & BLEU-4 & ARD & Tau & BLEU-4 & ARD & Tau & BLEU-4 & ARD & Tau & BLEU-4 & ARD & Tau \\
\midrule
XY-Cut~\cite{xycut} 
& 0.895 & 0.042 & 0.931 & 0.695 & 0.230 & 0.794 & 0.702 & 0.090 & 0.923 & 0.717 & 0.120 & 0.866 & 0.753 & 0.118 & 0.878 \\ 

LayoutReader~\cite{layoutreader,layoutlmv3} 
& \underline{0.988} & \textbf{0.004} & \underline{0.995} & 0.831 & 0.084 & 0.918 & 0.595 & 0.208 & 0.805 & 0.716 & 0.116 & 0.864 & 0.783 & 0.099 & 0.906 \\ 

MinerU~\cite{mineru} 
& 0.961 & \underline{0.025} & 0.969 & \underline{0.933} & \underline{0.037} & \underline{0.971} & \underline{0.923} & \underline{0.042} & \underline{0.965} & \underline{0.887} & \textbf{0.050} & \underline{0.932} & \underline{0.926} & \underline{0.039} & \underline{0.959} \\ 
\hline
\textbf{XY-Cut++(ours)} 
& \textbf{0.993} & \textbf{0.004} & \textbf{0.996} & \textbf{0.951} & \textbf{0.027} & \textbf{0.974} & \textbf{0.967} & \textbf{0.033} & \textbf{0.984} & \textbf{0.901} & \underline{0.064} & \textbf{0.942} & \textbf{0.953} & \textbf{0.037} & \textbf{0.972} \\ 
\bottomrule
\end{tabular}
\end{adjustbox}
\end{table*}

\begin{table*}[t]
\caption{Model Efficiency and Semantic Information Usage on DocBench-100 and OmniDocBench. \textit{Key Metrics:} FPS (Total Pages/Total Times) $\uparrow$. FPS values are averaged over 10 runs on Intel(R) Xeon(R) Gold 6326 CPU @ 2.90GHz with 256GB memory. Best results are in \textbf{bold}; second-best are \underline{underlined}. }
\label{tab:performance}
\centering
 \begin{adjustbox}{max width=\textwidth}
\begin{tabular}{@{}lcccl@{}} 
\toprule
Method & Semantic Info & \multicolumn{3}{c}{FPS} \\ 
\cmidrule{3-5} 
 &  & DocBench-100 & OmniDocBench & Mean \\
\midrule
XY-Cut~\cite{xycut} 
& \ding{55} & \underline{685} & \textbf{289} & \underline{487} \\
LayoutReader~\cite{layoutlmv3,layoutreader} 
& \ding{55} & 17 & 27 & 22 \\
MinerU~\cite{mineru} 
& \ding{55} & 10 & 12 & 11 \\
\midrule
\textbf{XY-Cut++(ours)} 
& \ding{51} & \textbf{781} & \underline{248} & \textbf{514} \\
\bottomrule
\end{tabular}
\end{adjustbox}
\end{table*}

In this section, we describe the experimental setup for evaluating our proposed method on the DocBench-100 benchmark. We compare our approach with several state-of-the-art baselines and demonstrate its effectiveness through both quantitative and qualitative analyses. 

\subsection{Dataset}
We evaluate on DocBench-100; detailed sources, fields, annotation pipeline, screening criteria, statistics, and usage protocols are described in Section~\ref{sec:docbench}. For fairness, we report both end-to-end and JSON-based protocols, and clarify FPS timing scope alongside Table~\ref{tab:performance}.

\subsection{Setup}

\paragraph{Baselines} We compare our method with the following state-of-the-art approaches:
\begin{itemize}[leftmargin=*,noitemsep]
% \begin{itemize}
    \item \textbf{XY-Cut}~\cite{xycut}: A classic method for projection-based segmentation. 
    \item \textbf{LayoutReader}~\cite{layoutlmv3, layoutreader}: A LayoutLMv3-based model fine-tuned on 500k samples. 
    \item \textbf{MinerU}~\cite{mineru}: An end-to-end document content extraction tool. 
\end{itemize}
\paragraph{Evaluation Metrics} We evaluate the performance using the following metrics:
\begin{itemize}[leftmargin=*,noitemsep]
% \begin{itemize}
    \item \textbf{BLEU-4}~\cite{bleu} ($\uparrow$): Measures the similarity between candidate and reference texts using up to 4-gram overlap. 
    \item \textbf{ARD} ($\downarrow$): Absolute Relative Difference, quantifies prediction accuracy by comparing predicted and actual values. 
    \item \textbf{Tau} ($\uparrow$): Kendall's Tau, measures the rank correlation between two sets of data. 
    \item \textbf{FPS} ($\uparrow$): Frames Per Second, a measure of how many frames a system can process per second. 
\end{itemize}

% Clarify block-level BLEU used in this work
\noindent\textbf{Block-level BLEU (detection order).} We evaluate the ordering of \emph{detection boxes} (block IDs), not textual content. Following the original BLEU definition~\cite{bleu}, we compute n-gram precisions on block identifiers and report BLEU-4 \emph{with} brevity penalty:
\begin{equation}
\mathrm{BLEU}\_4(\hat{s}, s) = BP \cdot \exp\left(\sum_{n=1}^4 \frac{1}{4} \log p_n\right),
\label{eq:bleu}
\end{equation}
where $p_n$ is the precision of block-level n-grams and $BP = \begin{cases}\exp(1 - r/c), & c \le r \\ 1, & c > r\end{cases}$ with $c$ the hypothesis length and $r$ the reference length. 

\noindent\textbf{FPS measurement scope.} Table~\ref{tab:performance} reports FPS for the ordering/sorting module only; upstream detection (e.g., PP-DocLayout) and downstream OCR/LM are excluded. Results are averaged over 10 runs on Intel(R) Xeon(R) Gold 6326 CPU @ 2.90GHz, 256GB RAM.

\subsection{Main Results}
\label{sec:main_results}

We evaluate our method on DocBench-100, analyzing component contributions and comparing against state-of-the-art baselines.  All metrics are computed over the union of $ D_c $ and $ D_r $, subsets unless specified. 

\subsubsection{Progressive Component Ablation} 
Table~\ref{tab:main} demonstrates the cumulative impact of each technical component:
\begin{itemize}[leftmargin=*,noitemsep]
% \begin{itemize}
\item \textbf{XY-Cut Baseline}: Achieves 0.749 BLEU-4 on complex layouts ($D_c$), showing limitations in handling complex layouts (e.g., L-shaped) elements. 

    \item \textbf{+Pre-Mask}: Improves BLEU-4 by 6.9 points (from 0.749 to 0.818) on $D_c$ via adaptive thresholding (Eq.~\eqref{eq:threshold}, $\beta=1.3$), reducing false splits by 15.9\%. 

    \item \textbf{+MGS}: Delivers a 19.7 absolute BLEU-4 gain on $D_c$ through three-phase segmentation, with density-aware splitting (Eq.~\eqref{eq:split}) reducing ARD by 29.7\%. 

    \item \textbf{+CMM}: Achieves near-perfect alignment (0.995 $\tau$) on $D_c$ through geometric constraints (Eq.~\eqref{eq:phi1}--\eqref{eq:phi4}), finalizing a 90.1\% ARD reduction from baseline. 
\end{itemize}

The complete model reduces ARD by 93.5\% compared to baseline (0.139 $\rightarrow$ 0.009), demonstrating superior ranking consistency. Notably, our method maintains balanced performance across both subsets (0.988 $\mu$-BLEU), proving effective for diverse layout types. 

\subsubsection{Architectural Analysis} 
We perform a systematic dissection of core components through controlled ablations to evaluate their contributions:

\textbf{Pre-Mask Processing (Pre-Mask):} To alleviate the "L-shaped" problem, we applied preliminary masking on highly dynamic elements such as titles, tables, and figures (see Methods Section~\ref{subsec:premask}). This approach reduced visual noise and improved reading order recovery, resulting in a 2.5-point BLEU-4 score increase, as shown in Table~\ref{tab:mgs_cmm}. 

\textbf{Multi-Granularity Segmentation (MGS):} Mask cross-layout is a very direct method to solve the problem that XY-Cut cannot segment L-shaped input. Pre-Cut realizes preliminary sub-page division through page analysis, thereby avoiding page content mixing affecting sorting. The adaptive splitting strategy enables reasonable segmentation through real-time density estimation. Table~\ref{tab:mgs_cmm} shows additive benefits of the mask cross-layout (+8.3), Pre-Cut (+9.2), and the adaptive splitting scheme (+10.1) over the baseline (0.822 BLEU-4). 

\textbf{Cross-Modal Matching (CMM):} As shown in Table~\ref{tab:mgs_cmm}, the single-stage strategy performs comparably to the baseline approach. This suggests that employing a detection model equipped with text, title, and annotation labels is sufficient to achieve nearly consistent performance across various scenarios. Notably, on the OmniDocBench dataset, which has few label categories, our method still achieves state-of-the-art results. Furthermore, we observe that the edge-weighted margin distance plays a crucial role among the four distance metrics examined. This finding highlights the significance of dynamic weights based on shallow semantics. In contrast, implementing cross-modal matching results in an overall improvement of 2.7 points in BLEU-4 scores. 

\subsubsection{Benchmark Comparison}
As shown in Table~\ref{tab:all}, our approach establishes new benchmarks across all evaluation dimensions. Specifically, it outperforms XY-Cut by a significant margin, achieving a +23.7 absolute improvement on \( D_c \) (from 74.9 to 98.6). Additionally, it surpasses LayoutReader by +5.3 on \( D_r \) (from 94.6 to 98.9), despite not using any learning-based components. Furthermore, our method achieves a Kendall's \( \tau \) of 0.996 overall, indicating near-perfect ordinal consistency (with \( p < 0.001 \) in the Wilcoxon signed-rank test). Visual results presented in Figure~\ref{fig:complex} and \ref{fig:simple} further demonstrate the robustness in handling multi-column layouts and cross-page elements, where previous methods frequently fail. 

To further validate the versatility and robustness of our proposed method, we conducted extensive evaluations on the OmniDocBench dataset~\cite{omnidocbench}, which features a diverse and challenging set of document images. As shown in Table~\ref{tab:omnidocbench}, our proposed method (XY-Cut++) achieves state-of-the-art (SOTA) performance across almost all layout types, despite challenging subpage nesting patterns (see Limitations). Notably, as shown in Table~\ref{tab:performance}, XY-Cut++ achieves a superior balance between performance and speed, attaining an average FPS of 514 across DocBench-100 and OmniDocBench. This performance surpasses even the direct projection-based XY-Cut algorithm, which achieves an average FPS of 487. The significant speed improvement of XY-Cut++ is primarily attributed to semantic filtering, which minimizes redundant processing by handling each block only once. In contrast, XY-Cut requires repeatedly partitioning blocks into different subsets, resulting in increased recursive depth and computational overhead. This optimization enhances computational efficiency without losing performance, making XY-Cut++ more robust and versatile for diverse document layouts. 

\section{Discussion}\label{sec6}
XY-Cut++ closes the gap between classical projection-based methods and neural models for block-level reading-order recovery through a simple three-stage pipeline. Directional ordering (left-to-right, top-to-bottom) is largely solved; the remaining pain points are fine-grained semantic segmentation (e.g., ambiguous block boundaries and caption linkage; see Figure~\ref{fig:app-badcases}) and sub-page structures. Across DocBench-100 and OmniDocBench, XY-Cut++ delivers large, consistent gains (Table~\ref{tab:main}, Table~\ref{tab:omnidocbench}) while retaining high throughput on CPU (Table~\ref{tab:performance}). On $D_c$, BLEU-4 rises from 0.749 to 0.986 and ARD falls from 0.233 to 0.023, with visual results (Figures~\ref{fig:complex},\ref{fig:simple}) confirming robustness on multi-column and L-shaped cases.

Mechanistically, the three components are complementary: Pre-Mask suppresses dominant elements to expose a stable backbone; MGS uses an adaptive density $\tau_d$ (Eq.~\eqref{eq:tau-density}) to choose the split axis (Eq.~\eqref{eq:split}); and CMM applies four geometric constraints (Eqs.~\eqref{eq:phi1}--\eqref{eq:phi4}) with scale-aware (Eq.~\eqref{eq:dyn-weights}) and semantic-specific weights (Eq.~\ref{eq:edge-weights}). Ablations indicate edge-weighted margins are especially effective.

Practically, reliable block-level ordering benefits RAG/LLM preprocessing by improving chunking and caption attachment, and XY-Cut++ is easy to deploy after a detector with coarse labels (e.g., PP-DocLayout). DocBench-100 helps standardize block-level evaluation and emphasizes difficult page topologies.

Limitations primarily stem from label granularity for fine-grained segmentation and from nested “sub-pages” (Figure~\ref{fig:app-badcases}); detector quality also matters. Heuristic hyperparameters (e.g., $\beta$ in Eq.~\eqref{eq:threshold}, $\theta_v$ in Eq.~\eqref{eq:split}) may require domain tuning. Future work targets: (i) sub-page detection with hierarchical reasoning to localize ordering, (ii) learning the split policy and edge weights from weak supervision to improve within-block segmentation, and (iii) coupling lightweight language priors (caption/title cues) with end-to-end RAG/LLM evaluation.

\section{Conclusion}\label{sec7}
We studied block-level reading-order recovery for real-world documents, a prerequisite for reliable RAG/LLM pipelines. We proposed XY-Cut++, a hierarchical, geometry-aware framework that integrates pre-mask processing, multi-granularity segmentation, and cross-modal matching. XY-Cut++ delivers precise block ordering and achieves state-of-the-art performance on DocBench-100 and OmniDocBench, while maintaining high throughput on CPU (cf. Tables~\ref{tab:main},~\ref{tab:performance}).

This accuracy--efficiency balance is achieved via two design choices: (i) shallow semantic labels used as structural priors and (ii) a hierarchical mask mechanism that stabilizes XY-Cut through density-aware splitting and edge-weighted matching. These components jointly improve accuracy without sacrificing throughput, making XY-Cut++ a practical module for production-grade document parsing. Beyond the algorithm, DocBench-100 offers a focused, block-level benchmark that emphasizes challenging page topologies and promotes standardized evaluation. Future work will extend XY-Cut++ with sub-page detection and lightweight language priors to better handle nested and fine-grained structures (see Section~\ref{sec6}).

\clearpage

\begin{appendices}
\section{Additional Visual Results}
\label{sec:appendix}

\subsection{Other Cases}
\begin{figure}[H]
    \centering
    \includegraphics[width=0.92\linewidth]{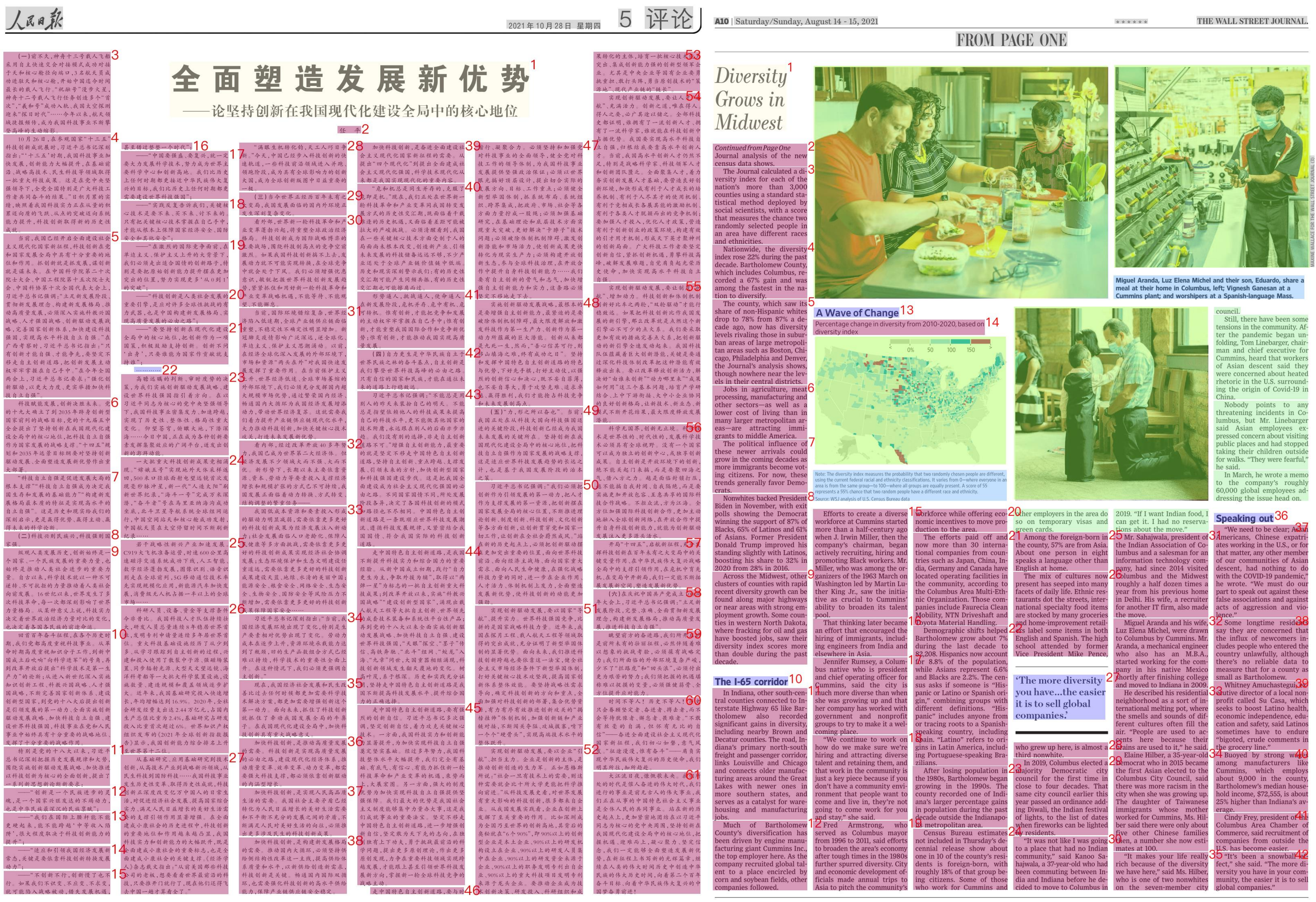}
    \caption{DocBench-100 (complex subset) --- Example A. A challenging multi-column page with spanning titles and interleaved elements. XY-Cut++ maintains a coherent block-level reading order across columns by combining pre-mask and density-aware splitting.}
    \label{fig:app-docbench-complex-1}
\end{figure}

\begin{figure}[H]
    \centering
    \includegraphics[width=0.92\linewidth]{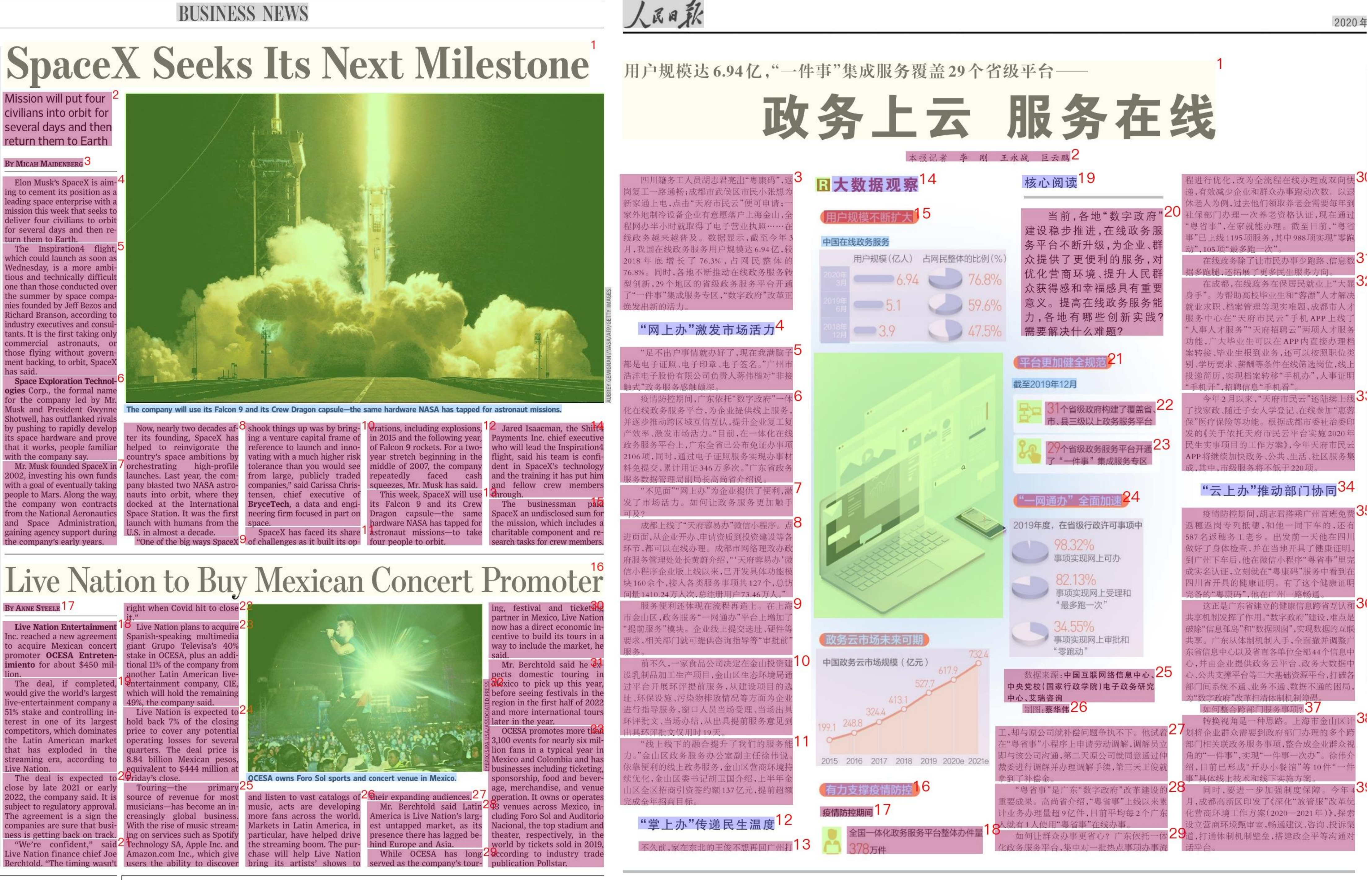}
    \caption{DocBench-100 (complex subset) --- Example B. Nested regions and interleaved figures/captions. The mask-then-remap strategy reduces L-shape artifacts and preserves local grouping for globally consistent ordering.}
    \label{fig:app-docbench-complex-2}
\end{figure}

\begin{figure}[H]
    \centering
    \includegraphics[width=1\linewidth]{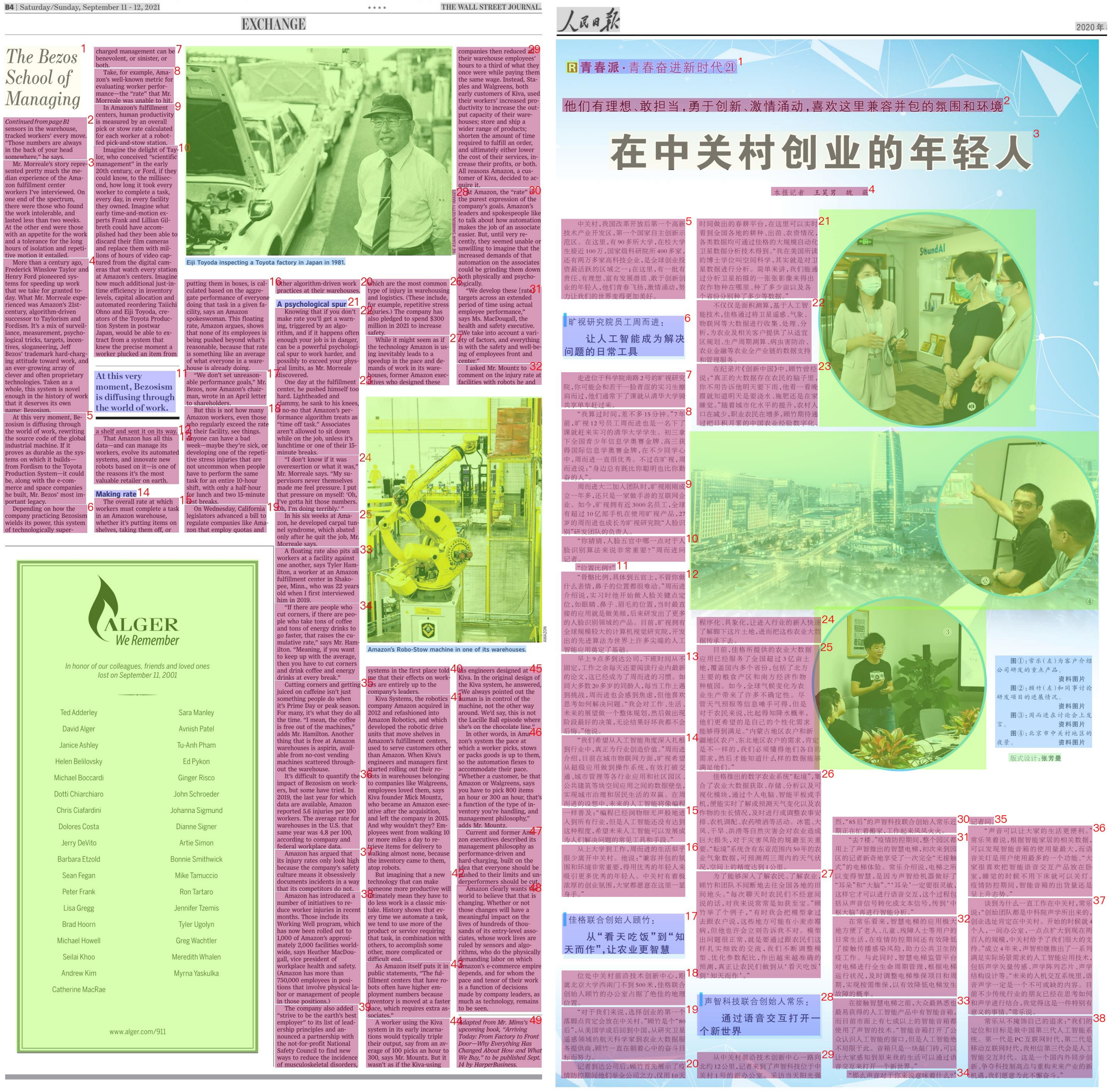}
    \caption{DocBench-100 (complex subset) --- Example C. L-shaped text around graphics. Pre-mask normalization avoids early incorrect splits, and the final sequence aligns with GT ordering.}
    \label{fig:app-docbench-complex-3}
\end{figure}

\begin{figure}[H]
    \centering
    \includegraphics[width=1\linewidth]{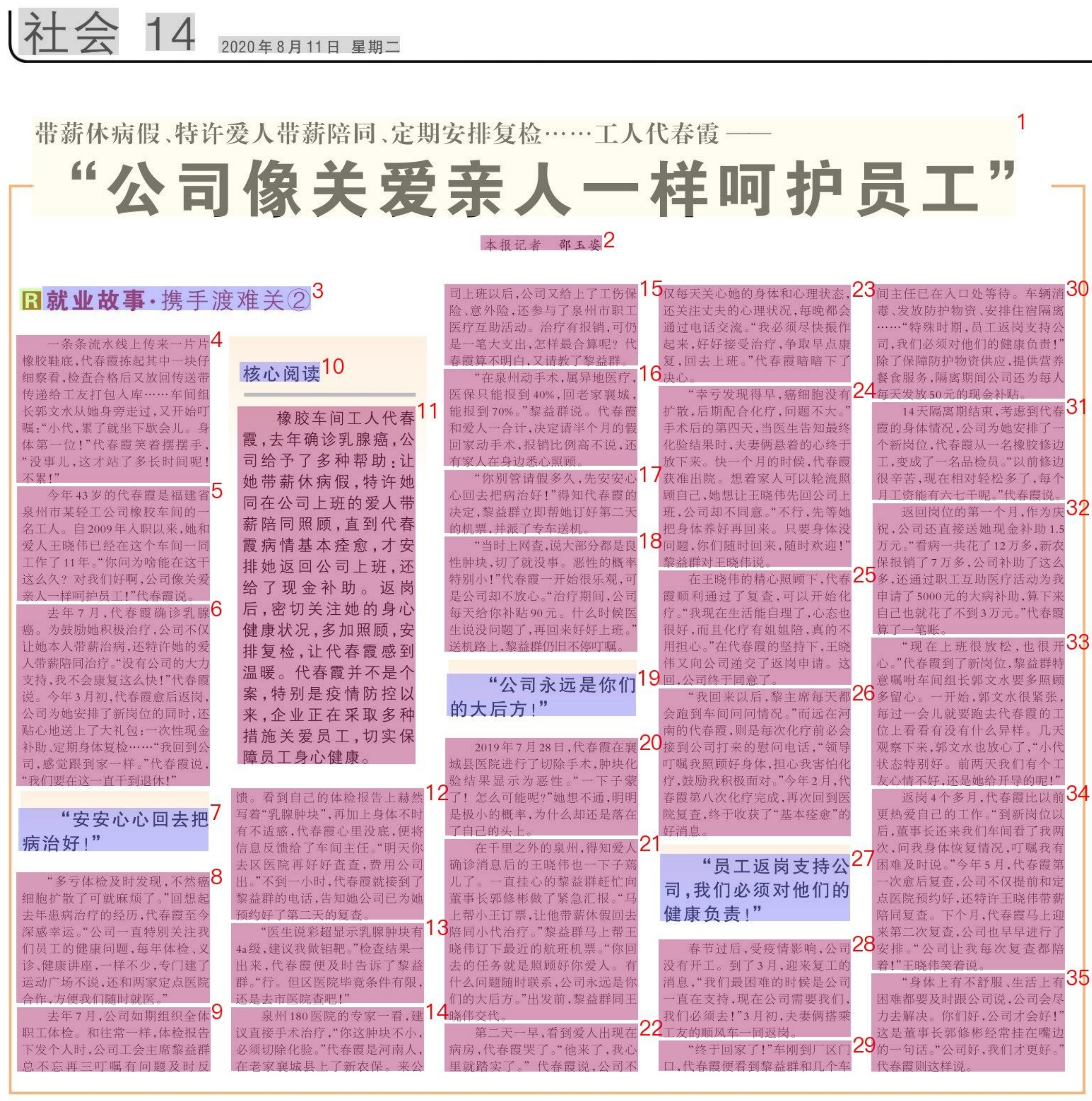}
    \caption{DocBench-100 (complex subset) --- Example D. Newspaper-like dense columns. The adaptive axis selection (horizontal vs. vertical) driven by regional density yields stable column-wise ordering.}
    \label{fig:app-docbench-complex-4}
\end{figure}

\begin{figure}[H]
    \centering
    \includegraphics[width=1\linewidth]{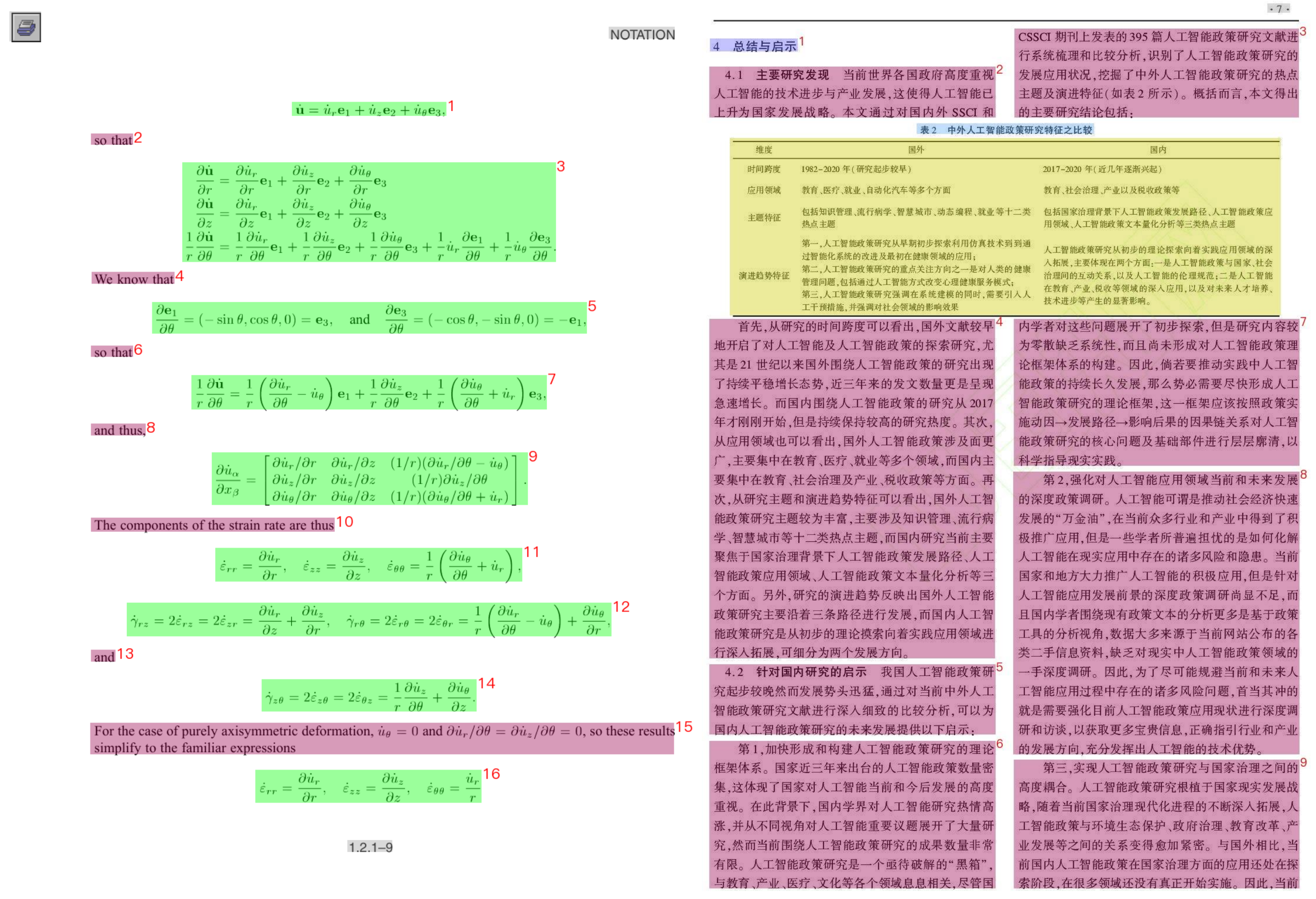}
    \caption{DocBench-100 (regular subset) --- Example A. A standard single/double-column page illustrating the method's stability on common business and academic layouts.}
    \label{fig:app-docbench-regular-1}
\end{figure}

\begin{figure}[H]
    \centering
    \includegraphics[width=1\linewidth]{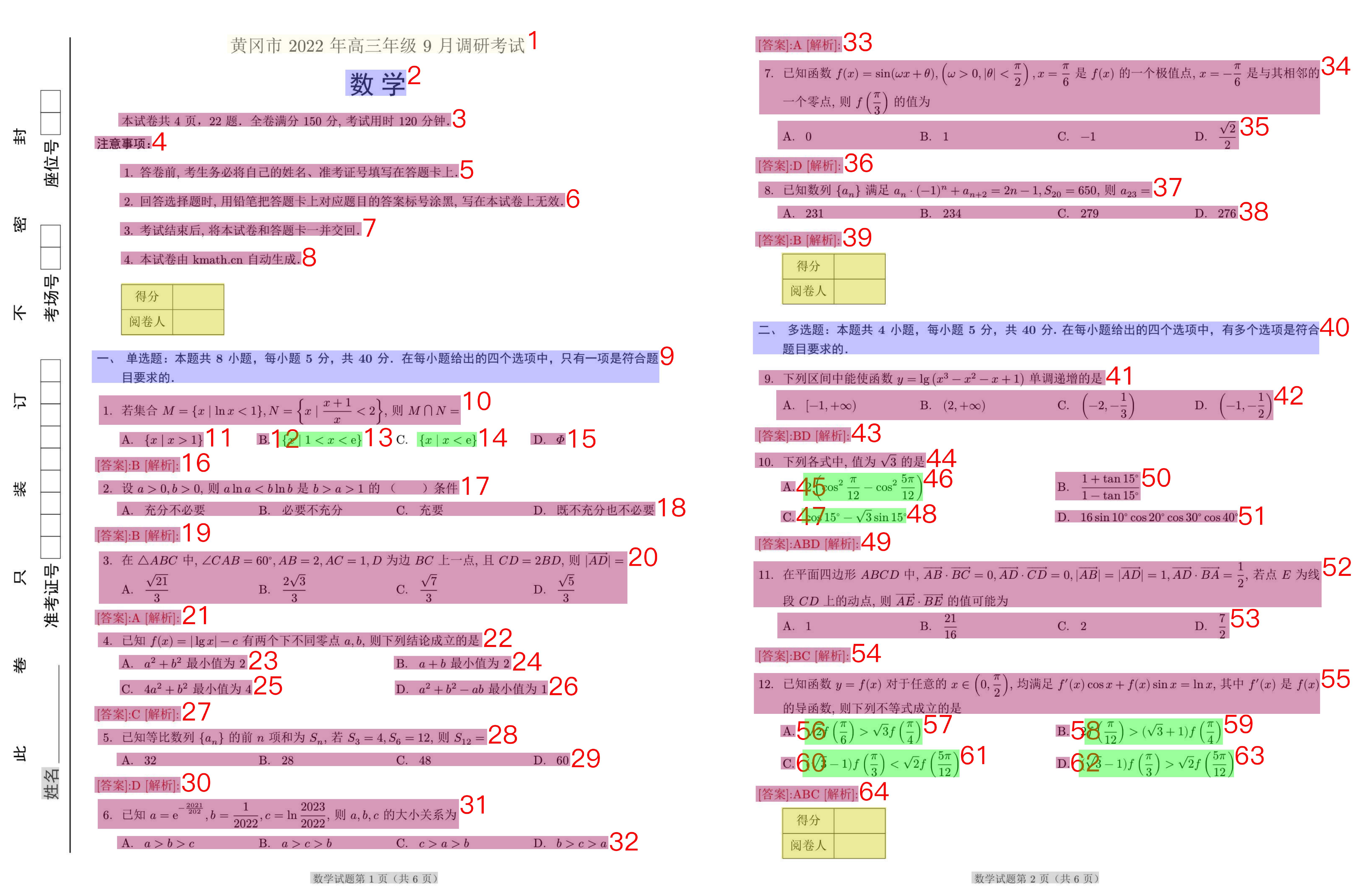}
    \caption{OmniDocBench (complex) --- Example A. A multi-column page with captions and side notes; geometry-aware matching preserves inter-column flow and caption linkage, illustrating robustness on challenging document layouts.}
    \label{fig:app-omni-double-1}
\end{figure}

\begin{figure}[H]
    \centering
    \includegraphics[width=0.95\linewidth]{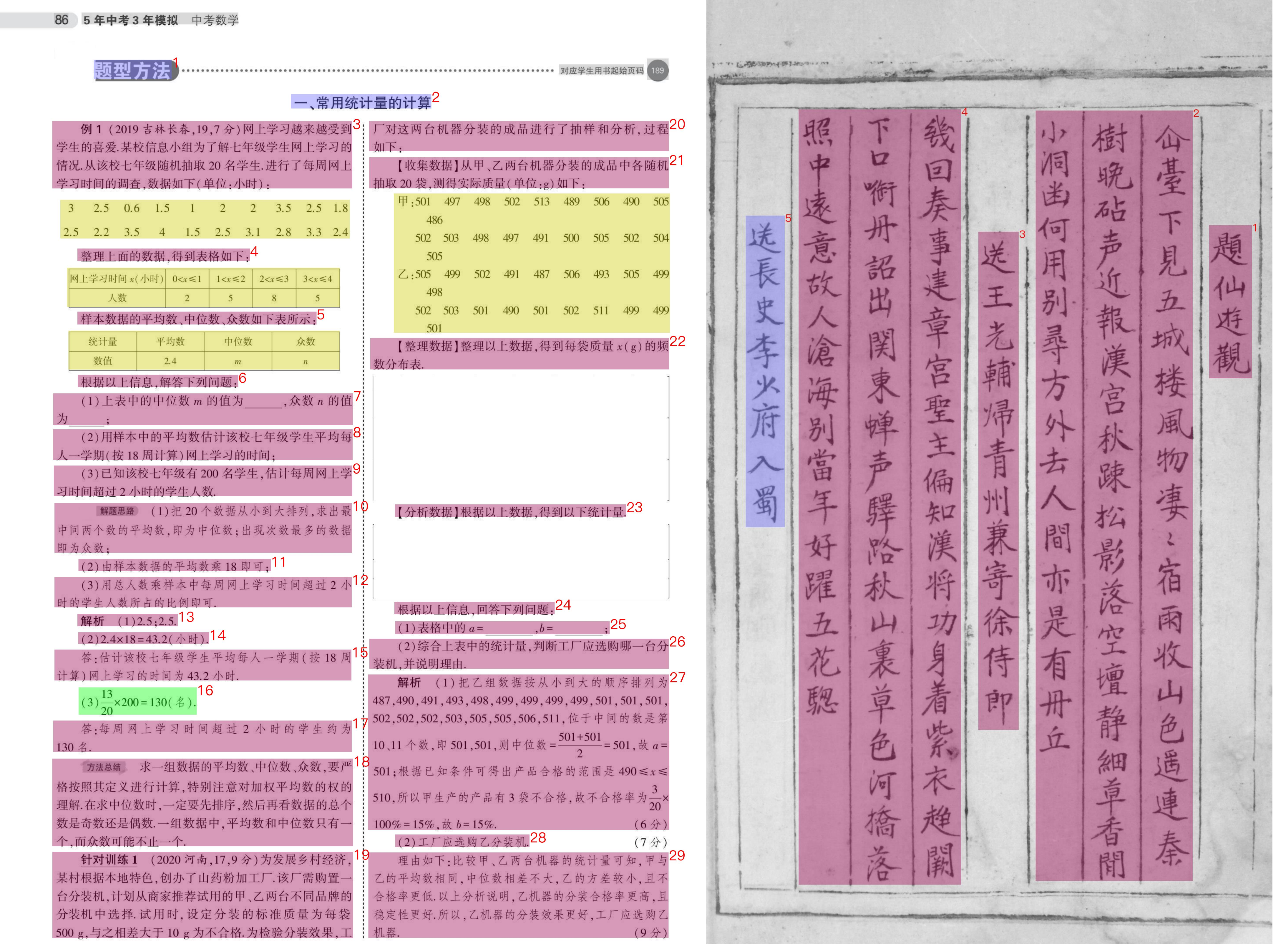}
    \caption{OmniDocBench (double) --- Example B and ancient-text case. Despite right-to-left script conventions and sparse annotations, the method maintains correct block-level order, evidencing robustness to atypical typography.}
    \label{fig:app-omni-double-2-ancient}
\end{figure}

\subsection{Failure Cases}
\begin{figure}[H]
    \centering
    \includegraphics[width=0.95\linewidth]{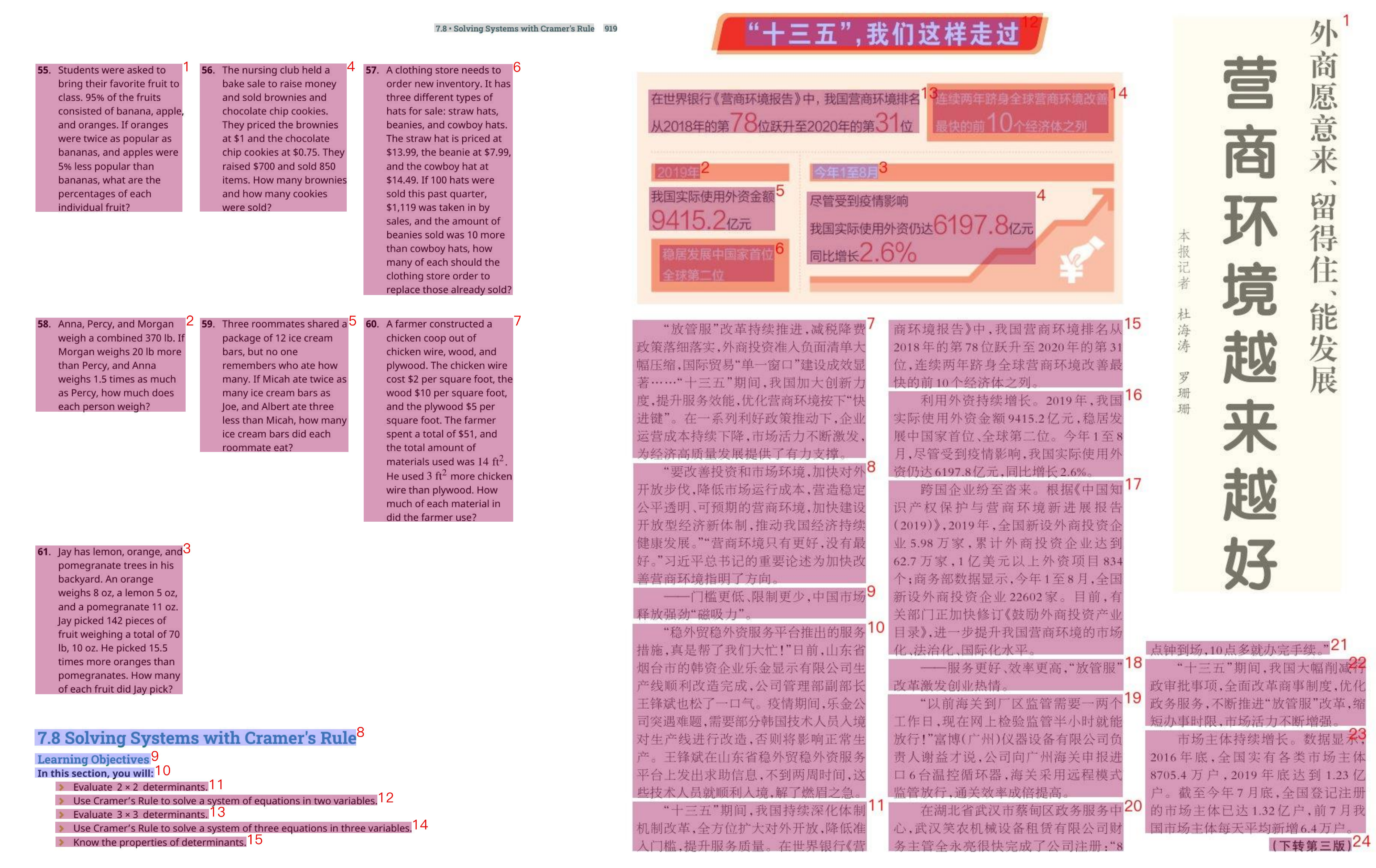}
    \caption{Challenging cases on OmniDocBench and DocBench-100. Failures stem from (1) insufficient fine-grained semantics and (2) sub-page complexity, leading to sorting errors; these motivate sub-page detection and stronger semantic priors.}
    \label{fig:app-badcases}
\end{figure}

\end{appendices}

% For Nature Portfolio eJP submissions, paste the .bbl contents here and remove the \bibliography command per journal instructions.
\bibliography{sn-bibliography}

\end{document}